\documentclass[lettersize,journal]{IEEEtran}
\usepackage[T1]{fontenc}
\usepackage{amsmath,amsfonts,amssymb,bm,booktabs,multirow,siunitx}
\usepackage{algpseudocode}
\usepackage{algorithm}
\usepackage{array}
\usepackage[caption=false,font=normalsize,labelfont=sf,textfont=sf]{subfig}
\usepackage{textcomp}
\usepackage{stfloats}
\usepackage{url}
\usepackage{verbatim}
\usepackage{graphicx}
\usepackage{xcolor}
\usepackage{cite}
\usepackage[colorlinks=true,citecolor=blue,linkcolor=black,urlcolor=black]{hyperref}
\newcommand{\std}[1]{\ensuremath{\,{\scriptstyle #1}}}

\newcolumntype{M}{S[table-format=2.2]}
\newcolumntype{U}{>{\scriptsize}S[table-format=2.2]}
\newcommand{\spm}{\std{\pm}}
\newcommand{\stdpm}[1]{\std{\pm\,#1}}
\hypersetup{pdfauthor={Jiawei Chen, Wenjun Wang, Wei Yu, Qiyao Peng},pdftitle={Dynamic Memory-enhanced Pattern Relevance Prediction for Long-tailed Graph Classification}}
\begin{document}
\title{Dynamic Memory-enhanced Pattern Relevance Prediction for Long-tailed Graph Classification}
\author{Jiawei~Chen, Wenjun~Wang, Wei~Yu, and Qiyao~Peng%
\thanks{Jiawei Chen, Wenjun Wang, and Qiyao Peng are with Tianjin University, Tianjin, China (e-mail: ppw@tju.edu.cn, wjwang@tju.edu.cn, qypeng@tju.edu.cn).}%
\thanks{Wei Yu is with Zhejiang Yuexiu University, Shaoxing, China (e-mail: weiyu@tju.edu.cn).
\textit{Corresponding author: Qiyao Peng.}}}
\markboth{Preprint}{Chen et al.: Dynamic Memory-enhanced Pattern Relevance Prediction for Long-tailed Graph Classification}
\maketitle
\begin{abstract}
Long-tailed graph classification (LTGC) is common in real-world graph datasets, where a few large head graphs provide rich structural information while many smaller tail graphs offer fewer structural cues, making it harder to learn discriminative representations for tail graphs. To address this size-based imbalance, we propose Dynamic Memory-enhanced Pattern Relevance Prediction (DMPRP), which transfers structural patterns from head graphs to tail graphs. Given a graph representation as a query, a dynamic expert memory module inspired by Knowledge-diverse Experts (KDEX) generates a residual vector that augments the original representation. We train the module using co-occurrence triplets sampled from head graphs at the node, rooted subgraph, and random-walk neighborhood levels. Experiments on five graph classification benchmarks show that DMPRP improves tail performance, and ablation studies show that both the expert module and random-walk supervision contribute to these gains. A controlled experiment on PTC further examines pattern transfer when smaller graphs are denser than larger ones. Our code is available at \url{https://anonymous.4open.science/r/DMPRP-73A5}.
\end{abstract}
\section{Introduction}
\IEEEPARstart{G}{raph} classification is a fundamental problem in graph machine learning with broad applications~\cite{zeng2023substructure} such as molecular property prediction~\cite{wu2018moleculenet}, program/malware analysis~\cite{allamanis2017learning}, and social/community understanding~\cite{wu2021gnnsurvey,gilmer2017neural}. Despite the success of graph neural networks (GNNs) on balanced benchmarks~\cite{kipf2017semi,xu2019gin}, real-world graph datasets are often \emph{imbalanced}~\cite{ma2021comprehensive}, which can induce systematic bias toward dominant training regimes and deteriorate generalization on under-represented samples~\cite{liu2025imbalanced_survey,ma2025cilg_survey}.

\suppressfloats[t] %
\begin{figure}[t]
  \centering
  \includegraphics[width=0.96\linewidth]{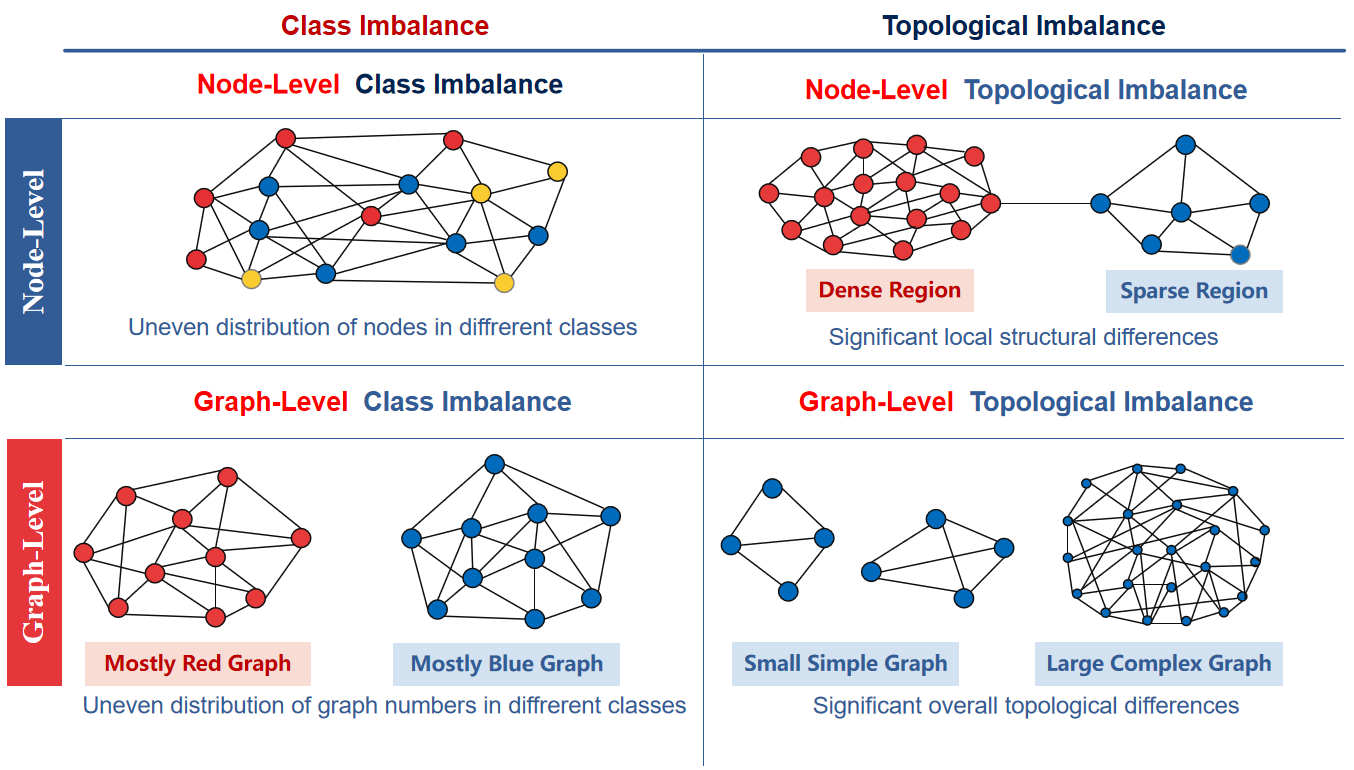}
  \caption{A taxonomy of imbalance settings in graph learning~\cite{qin2024iglbench}. Imbalance can be induced at the node or graph level, and it can stem from label frequency (\emph{class-imbalance}) or structural statistics (\emph{topology-imbalance}). In this work, we focus on \textbf{graph-level topology imbalance}, which induces long-tailed graph classification.}
  \label{fig:imbalance_manipulator}
\end{figure}

As summarized in Fig.~\ref{fig:imbalance_manipulator}, imbalance in graph learning can be categorized along two axes: \emph{(i) node-level vs.\ graph-level} and \emph{(ii) class-imbalance vs.\ topology-imbalance}~\cite{qin2024iglbench}. 
Most long-tailed learning studies emphasize \emph{class-imbalance}, where the number of training instances per class follows a long-tailed distribution and is typically addressed by re-weighting, margin modification, or decoupled training strategies~\cite{cui2019classbalanced,cao2019ldam,kangdecoupling}. 
In contrast, the long-tailed graph classification (LTGC) setting we study is primarily driven by \textbf{graph-level topology imbalance}: graphs exhibit a highly skewed distribution of structural statistics (e.g., graph size, density, and substructure richness), which yields \emph{structure-abundant} head graphs and \emph{structure-scarce} tail graphs~\cite{zhao2022topoimb,liu2022soltgnn}. 
This setting is challenging because the supervision signal is tightly coupled with topology: tail graphs not only have fewer samples but also provide less structural evidence, making them harder to represent faithfully and easier to overfit~\cite{chen2021renode}.

Graph-level topology imbalance breaks several common assumptions behind standard GNN training.
First, when graphs are small or sparse, message passing can have limited receptive-field diversity, making representations more sensitive to idiosyncratic structures and noise~\cite{liu2021tailgnn}. 
Second, training dominated by large head graphs can bias the decision boundary toward patterns correlated with dense structures, harming tail graphs systematically~\cite{zhao2022topoimb}. 
Finally, na\"ive re-balancing schemes designed for class-imbalance are often insufficient here, because the core difficulty is a \emph{structural capacity gap} between head and tail graphs rather than label frequency alone~\cite{wang2022g2gnn}.

To tackle these challenges, we propose \emph{Dynamic Memory-enhanced Pattern Relevance Prediction} (\textbf{DMPRP}), a size-oriented framework that learns transferable structural patterns from head graphs and uses them to complement tail graph representations~\cite{liu2022soltgnn}. 
Concretely, DMPRP performs pattern relevance prediction via a \textbf{dynamic expert memory} inspired by Knowledge-diverse Experts (KDEX)~\cite{mao2025kdex}, enabling query-conditioned retrieval and aggregation of complementary patterns. 
In addition, beyond conventional node-level and subgraph-level co-occurrence supervision, we introduce \textbf{neighborhood-level random-walk subgraph triplets} and unify all three granularities into a multi-granularity contrastive objective, strengthening local-to-global consistency under topology-scarce tail supervision~\cite{you2020graphcl,schroff2015facenet}.

Our main contributions are:
\begin{itemize}
  \item We formalize LTGC from the perspective of \textbf{graph-level topology imbalance} (Fig.~\ref{fig:imbalance_manipulator}) and highlight why it differs from standard class-imbalanced learning~\cite{qin2024iglbench,liu2025imbalanced_survey}.
  \item We propose \textbf{DMPRP}, featuring a \textbf{KDEX-style dynamic expert memory} for query-conditioned pattern relevance prediction and representation complement~\cite{mao2025kdex}.
  \item We introduce \textbf{neighborhood-level random walk triplet mining} and a unified multi-granularity contrastive objective to better exploit transferable patterns under limited tail supervision~\cite{you2020graphcl}.
  \item Experiments on multiple LTGC benchmarks demonstrate consistent improvements, especially on tail graphs, and ablations validate the effectiveness of each component.
\end{itemize}

Section~\ref{sec:related} reviews related work. Section~\ref{sec:prelim} defines graph-level topology imbalance and introduces the notation. Section~\ref{sec:model} describes how DMPRP learns complementary representations from head graphs. Section~\ref{sec:exp} evaluates classification performance and examines the contributions of the expert module and triplet supervision. Section~\ref{sec:Conclusion} concludes the paper.

\section{Related Work}
\label{sec:related}

\subsection{Imbalanced Learning on Graphs}
Imbalanced learning has been extensively studied in \emph{independent and identically distributed (i.i.d.)} settings, where imbalance primarily arises from long-tailed label distributions and is commonly addressed via class re-weighting~\cite{lin2017focal}, margin reshaping, or decoupled optimization~\cite{cui2019classbalanced,cao2019ldam,kangdecoupling}. 
For graph data, recent surveys and benchmarks summarize that imbalance can arise at both node and graph levels, and can be driven by either class frequency or structural statistics~\cite{liu2025imbalanced_survey,qin2024iglbench,ma2025cilg_survey}. 
Representative methods for imbalanced node learning include synthetic over-sampling and subgraph sampling strategies (e.g., GraphSMOTE and PC-GNN)~\cite{zhao2021graphsmote,liu2021pcgnn}. 
Beyond class frequency, \emph{topology-related} imbalance has been explicitly recognized as a key issue; ReNode proposes topology-aware re-weighting for semi-supervised node learning~\cite{chen2021renode}. 
Our work focuses on \textbf{graph-level topology imbalance}, where the key bottleneck is the scarcity of structural evidence in tail graphs.

\subsection{Long-tailed and Imbalanced Graph Classification}
Imbalanced graph classification has received increasing attention, but is comparatively less explored than node-level imbalance. 
G$^2$GNN addresses class-imbalanced graph classification by constructing a graph-of-graphs and propagating supervision globally~\cite{wang2022g2gnn}. 
For LTGC driven by \emph{topology/size skew}, SOLT-GNN presents a size-oriented formulation and shows that transferring reusable patterns from head graphs can complement tail representations~\cite{liu2022soltgnn}. 
SOLT-GNN learns node and subgraph co-occurrences in head graphs and adapts its shared pattern memory to each query through scaling and shifting. Other methods address graph size more directly. SizeShiftReg coarsens training graphs to simulate a size shift and regularizes the encoder across graph sizes~\cite{buffelli2022sizeshiftreg}. SIMBA links graphs with similar structural features and uses energy propagation to reweight training examples~\cite{qin2025simba}.

\subsection{Graph Contrastive Learning and Triplet Objectives}
Contrastive learning has become a powerful paradigm for representation learning, often instantiated by InfoNCE-style objectives~\cite{oord2018cpc} and extended to graph domains via augmentation-based frameworks such as GraphCL~\cite{you2020graphcl}. 
Triplet-based objectives are a classical form of metric/contrastive learning that enforce relative similarity constraints between a query, a positive, and a negative sample~\cite{schroff2015facenet}. 
In our setting, we unify node-, subgraph-, and neighborhood-level (random-walk) triplets into a multi-granularity relevance objective, providing complementary structural signals for topology-scarce tail graphs.

\subsection{Memory and Retrieval Mechanisms}
Memory-augmented models and retrieval-based mechanisms have been widely explored to enhance prediction by leveraging external or learned knowledge stores~\cite{miller2016keyvalue,graves2016dnc}. 
Recent work further shows that retrieving relevant information conditioned on the input/query can significantly improve robustness and generalization~\cite{guu2020realm}. 
KDEX addresses class-frequency imbalance by training experts for different class distributions and routing their predictions at test time~\cite{mao2025kdex}. In DMPRP, the router combines expert outputs into a residual vector that is added to the graph embedding before classification.

Table~\ref{tab:related_methods} compares these approaches. DMPRP builds on SOLT-GNN's use of head-graph co-occurrence and extends its training signal to neighborhoods sampled by random walks. These walks expose the predictor to sequences of adjacent nodes, alongside individual nodes and rooted subgraphs. On D\&D, adding walk supervision to the node and subgraph objectives raises mean tail accuracy from 71.17\% to 75.64\% (Table~\ref{tab:ablation_5datasets}).

\begin{table}[!ht]
\centering
\caption{Learning settings and mechanisms for graph classification.}
\label{tab:related_methods}
\footnotesize
\setlength{\tabcolsep}{3pt}
\renewcommand{\arraystretch}{1.15}
\begin{tabular}{@{}l>{\raggedright\arraybackslash}p{0.22\columnwidth}>{\raggedright\arraybackslash}p{0.47\columnwidth}@{}}
\toprule
Method & Setting & Mechanism \\
\midrule
SOLT-GNN~\cite{liu2022soltgnn} & Size imbalance & Adapts shared memory to a query using node and subgraph co-occurrence. \\
SizeShiftReg~\cite{buffelli2022sizeshiftreg} & Size shift & Aligns representations of original and coarsened training graphs. \\
SIMBA~\cite{qin2025simba} & Size imbalance & Smooths features across graphs and reweights training examples. \\
KDEX~\cite{mao2025kdex} & Class imbalance & Routes predictions from experts trained for different class distributions. \\
\midrule
\textbf{DMPRP} & Size imbalance & Combines expert outputs into a residual using node, subgraph, and random-walk triplet supervision. \\
\bottomrule
\end{tabular}
\end{table}

\section{Preliminaries}
\label{sec:prelim}

\subsection{Problem Setting: Topology-imbalanced Long-tailed Graph Classification}
\label{sec:prelim:problem}
We study long-tailed graph classification (LTGC) under \emph{graph-level topology imbalance}, where graphs exhibit highly skewed structural statistics (e.g., graph size, density, and substructure richness).
Let $\mathcal{D}=\{(G_i, y_i)\}_{i=1}^{N}$ denote a labeled graph dataset, where $G_i=(V_i,E_i)$ and $y_i \in \{1,\dots,C\}$ is the class label.
Following size-oriented LTGC~\cite{liu2022soltgnn}, we use the number of nodes $|V_i|$ as a representative topology indicator.
The dataset is partitioned into \emph{head} graphs and \emph{tail} graphs according to $|V_i|$ (or more generally, topology statistics), where head graphs are structurally rich and tail graphs are structurally scarce.
Our goal is to learn a classifier that performs well on both head and tail graphs, especially improving generalization on topology-scarce tail graphs.

\subsection{GNN Encoder and Graph Representation}
\label{sec:prelim:encoder}
We adopt a message-passing GNN as the encoder to obtain node- and graph-level representations~\cite{kipf2017semi,xu2019gin}.
Given node features $\mathbf{X}_i$, the encoder produces node embeddings $\{\mathbf{h}_v\}_{v\in V_i}$ through $L$ layers of neighborhood aggregation.
A permutation-invariant readout function $\mathrm{READOUT}(\cdot)$ (e.g., sum/mean pooling) aggregates node embeddings to a graph representation:
\begin{equation}
\mathbf{h}_{G_i} = \mathrm{READOUT}\big(\{\mathbf{h}_v\}_{v\in V_i}\big).
\end{equation}

\begin{figure*}[t]
  \centering
  \resizebox{\textwidth}{!}{%
    \setlength{\unitlength}{1pt}%
    \begin{picture}(1818,492)
      \put(0,0){\includegraphics[width=1818pt,height=492pt]{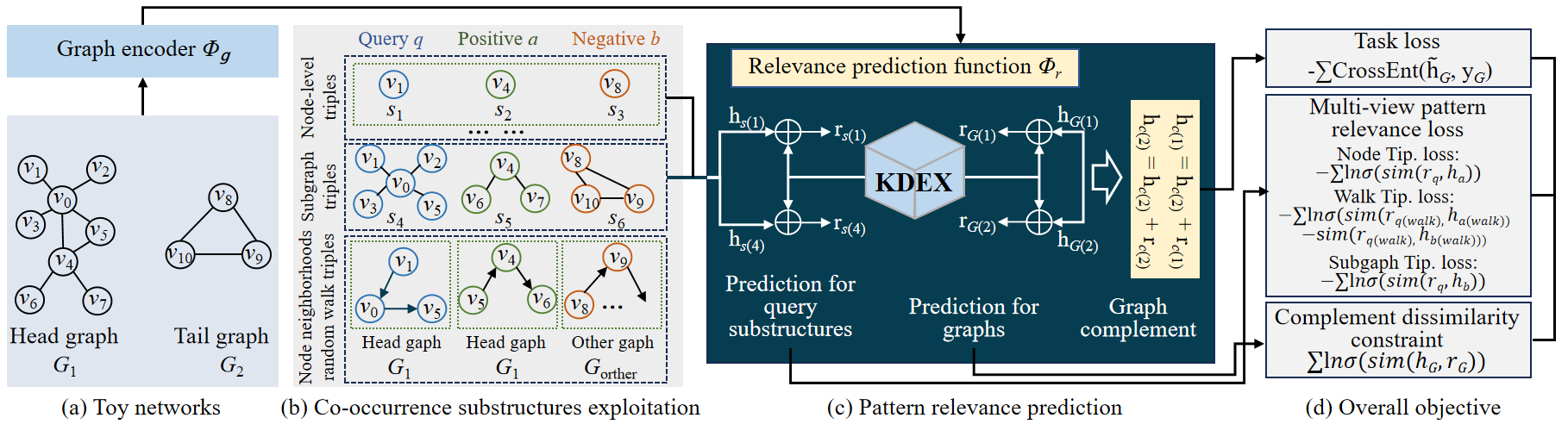}}
      \put(1467,40){{\color{white}\rule{309pt}{442pt}}}
      \put(1467,408){{\color[RGB]{23,44,81}\rule{309pt}{75pt}}}
      \put(1469,410){{\color[RGB]{242,242,242}\rule{305pt}{71pt}}}
      \put(1467,142){{\color[RGB]{23,44,81}\rule{309pt}{256pt}}}
      \put(1469,144){{\color[RGB]{242,242,242}\rule{305pt}{252pt}}}
      \put(1467,40){{\color[RGB]{23,44,81}\rule{309pt}{92pt}}}
      \put(1469,42){{\color[RGB]{242,242,242}\rule{305pt}{88pt}}}
      \put(1469,455){\includegraphics[viewport={1101.612bp 324.709bp 1330.334bp 342.707bp},clip,width=305pt,height=24pt]{DMPRP.png}}
      \put(1469,415){\includegraphics[viewport={1101.612bp 294.713bp 1330.334bp 322.460bp},clip,width=305pt,height=37pt]{DMPRP.png}}
      \put(1469,342){\includegraphics[viewport={1101.612bp 248.219bp 1330.334bp 285.714bp},clip,width=305pt,height=50pt]{DMPRP.png}}
      \put(1469,286){\includegraphics[viewport={1101.612bp 206.224bp 1330.334bp 244.469bp},clip,width=305pt,height=51pt]{DMPRP.png}}
      \put(1469,209){\includegraphics[viewport={1101.612bp 152.231bp 1330.334bp 206.224bp},clip,width=305pt,height=72pt]{DMPRP.png}}
      \put(1469,147){\includegraphics[viewport={1101.612bp 113.236bp 1330.334bp 149.981bp},clip,width=305pt,height=49pt]{DMPRP.png}}
      \put(1469,77){\includegraphics[viewport={1101.612bp 65.242bp 1330.334bp 104.987bp},clip,width=305pt,height=53pt]{DMPRP.png}}
      \put(1469,45){\includegraphics[viewport={1101.612bp 41.995bp 1330.334bp 65.242bp},clip,width=305pt,height=31pt]{DMPRP.png}}
    \end{picture}%
  }
  \caption{Overall framework of \textbf{DMPRP}. (a) A toy example illustrating head and tail graphs under graph-level topology imbalance. (b) Multi-granularity co-occurrence triplets at node, subgraph, and neighborhood random-walk levels. (c) KDEX-based pattern relevance prediction function $\phi_r$ produces query-conditioned complements for substructures and graphs. (d) Overall objective combining task loss, multi-view pattern relevance loss, and complement dissimilarity constraint.}
  \label{fig:framework}
\end{figure*}

\subsection{Multi-granularity Substructures}
\label{sec:prelim:substructures}
To capture transferable structural patterns at different scales, we consider three granularities of substructures~\cite{bouritsas2022improving}:
\textbf{(i) node-level}, \textbf{(ii) subgraph-level}, and \textbf{(iii) neighborhood-level random-walk subgraphs}.
For a node $v\in V_i$, its node-level embedding is $\mathbf{h}_v$.
For subgraph-level patterns, we extract a rooted subgraph $S_i(v)$ (e.g., via BFS expansion) and compute its representation by pooling the embeddings of nodes inside the subgraph:
\begin{equation}
\mathbf{h}_{S_i(v)} = \mathrm{READOUT}\big(\{\mathbf{h}_u\}_{u\in V(S_i(v))}\big).
\end{equation}
For neighborhood-level random-walk subgraphs, we perform a random walk of length $s$ starting from $v$ to obtain a visited node set $V^{rw}_i(v)$, and form the induced subgraph $R_i(v)=G_i[V^{rw}_i(v)]$.
Its representation is similarly computed as
\begin{equation}
\mathbf{h}_{R_i(v)} = \mathrm{READOUT}\big(\{\mathbf{h}_u\}_{u\in V^{rw}_i(v)}\big).
\end{equation}

\subsection{Triplet-based Contrastive Objective}
\label{sec:prelim:triplet}
We adopt triplet-based contrastive learning to model co-occurrence relations between substructures.
A triplet is denoted as $(q,a,b)$, where $q$ is a query, $a$ is a positive sample, and $b$ is a negative sample.
Given representations $\mathbf{z}_q$, $\mathbf{z}_a$, and $\mathbf{z}_b$, we measure similarity by $\mathrm{sim}(\cdot,\cdot)$ (cosine similarity) and optimize a ranking-style triplet objective:
\begin{equation}
\mathcal{L}_{\mathrm{tri}} = - \log \sigma\Big( \mathrm{sim}(\mathbf{z}_q,\mathbf{z}_a) - \mathrm{sim}(\mathbf{z}_q,\mathbf{z}_b) \Big),
\end{equation}
where $\sigma(\cdot)$ is the sigmoid function.
This objective encourages the query to be closer to the positive than the negative in the embedding space, and can be viewed as a classical metric learning formulation~\cite{schroff2015facenet}.

\subsection{Pattern Complement Interface}
\label{sec:prelim:complement}
We define a generic \emph{pattern complement} interface used throughout the paper.
Given a representation $\mathbf{h}$ (node/substructure/graph), a relevance predictor $\phi_r(\cdot)$ outputs a complementary vector $\mathbf{r}=\phi_r(\mathbf{h})$ that injects transferable pattern evidence.
The complemented representation is defined as
\begin{equation}
\tilde{\mathbf{h}} = \mathbf{h} + \mathbf{r}.
\end{equation}
In the next section, we instantiate $\phi_r(\cdot)$ with a KDEX-style dynamic expert memory and design multi-granularity triplets to learn relevance-aware complements for topology-scarce tail graphs.

\section{The Proposed Model}
\label{sec:model}

We propose \emph{Dynamic Memory-enhanced Pattern Relevance Prediction} (\textbf{DMPRP}) to address long-tailed graph classification under \textbf{graph-level topology imbalance} (Sec.~\ref{sec:prelim:problem}).
As shown in Fig.~\ref{fig:framework}, DMPRP follows a size-oriented pattern transfer paradigm: it learns transferable structural patterns from topology-rich head graphs and complements topology-scarce tail graphs through \emph{pattern relevance prediction}.
DMPRP consists of three key components:
(i) a GNN encoder $\phi_g$ that produces node/substructure/graph representations (Sec.~\ref{sec:prelim:encoder});
(ii) a KDEX-style \emph{dynamic expert memory} that performs query-conditioned relevance prediction and outputs complementary vectors (Sec.~\ref{sec:model:kdex});
and (iii) multi-granularity co-occurrence triplets (node/subgraph/random-walk neighborhood) that provide relevance-aware supervision for mining transferable patterns (Sec.~\ref{sec:model:triplets}).
These components are jointly optimized with a classification objective, a multi-view relevance loss, and a complement dissimilarity regularizer (Sec.~\ref{sec:model:objective}).

\subsection{KDEX-style Dynamic Expert Memory for Pattern Relevance Prediction}
\label{sec:model:kdex}
DMPRP uses a dynamic expert memory to predict a complement for each query.
Given an input representation $\mathbf{h}$ (node/substructure/graph), the relevance prediction function $\phi_r(\cdot)$ outputs a complement $\mathbf{r}=\phi_r(\mathbf{h})$, and we obtain the complemented representation by $\tilde{\mathbf{h}}=\mathbf{h}+\mathbf{r}$ (Sec.~\ref{sec:prelim:complement}).
Compared with static matching, a dynamic expert memory enables query-conditioned composition of diverse transferable cues, which is particularly beneficial when tail graphs are topology-scarce.

\paragraph{Expert bank}
We maintain a set of $M$ learnable experts $\mathcal{E}=\{E_m(\cdot)\}_{m=1}^{M}$, where each expert maps the input to a candidate complement:
\begin{equation}
\mathbf{r}_m = E_m(\mathbf{h}), \quad m=1,\dots,M .
\label{eq:expert_out}
\end{equation}
Each expert uses the query to scale and shift a learned $p\times p$ matrix, where $p$ is the expert width. It then projects each row to the embedding dimension $d$, applies a leaky ReLU, and takes a learned weighted sum of the rows to produce $\mathbf{r}_m$.

\paragraph{Query-conditioned gating}
A gating network~\cite{shazeer2017outrageously} produces expert weights conditioned on the query:
\begin{equation}
\boldsymbol{\alpha}=\mathrm{softmax}\!\left(g(\mathbf h)\right)\in\mathbb{R}^{M},
\label{eq:gating}
\end{equation}
where $g$ is a two-layer MLP with a ReLU hidden layer of width $p$ and $M$ output scores. For each query, we evaluate all $M$ experts.

\paragraph{Dynamic composition}
The final complement is obtained by dynamically composing the expert outputs:
\begin{equation}
\mathbf{r}=\phi_r(\mathbf{h})=\sum_{m=1}^{M}\alpha_m \mathbf{r}_m.
\label{eq:kdex_compose}
\end{equation}
We apply the same relevance function to both (i) query substructures (node/subgraph/random-walk neighborhood) and (ii) graph-level embeddings, yielding $\mathbf{r}_q$ and $\mathbf{r}_G$, respectively (Fig.~\ref{fig:framework}(c)).

\subsection{Multi-granularity Co-occurrence Triplet Mining}
\label{sec:model:triplets}
To exploit transferable co-occurrence patterns, we construct triplets at three granularities: node-level, subgraph-level, and neighborhood-level random-walk subgraphs.
Each triplet is denoted as $(q,a,b)$ with query $q$, positive $a$, and negative $b$ (Sec.~\ref{sec:prelim:triplet}).
All triplets are sampled from topology-rich graphs (i.e., head graphs) to provide stronger structural evidence for learning transferable patterns, and are used to regularize relevance prediction (Fig.~\ref{fig:framework}(b)).

\paragraph{Node-level triplets}
For each graph $G_i$, we sample two nodes $v,u\in V_i$ ($v\neq u$) and treat them as a positive pair.
We sample a node $u'\in V_j$ from another graph $G_j$ ($j\neq i$) as negative.
The triplet is $q=v$, $a=u$, and $b=u'$, and we denote the set by $\mathcal{T}_{node}$.

\paragraph{Subgraph-level triplets}
For each node $v\in V_i$, we extract a rooted subgraph $S_i(v)$ (e.g., BFS expansion) and use it as the query $q$.
A positive subgraph $a$ is sampled from the same graph (e.g., a rooted subgraph around another node), while a negative subgraph $b$ is sampled from a different graph $G_j$.
We denote the set by $\mathcal{T}_{sub}$.

\paragraph{Neighborhood-level random-walk triplets}
To capture message-passing-consistent neighborhood patterns, we introduce random-walk subgraph triplets.
For a node $v\in V_i$, we generate the random-walk induced subgraph $R_i(v)$ (Sec.~\ref{sec:prelim:substructures}) and use it as the query $q$.
The positive $a=R_i(u)$ is generated from another node $u\neq v$ in the same graph $G_i$, and the negative $b=R_j(u')$ is generated from a node $u'$ in a different graph $G_j$.
We denote the set by $\mathcal{T}_{rw}$.

\paragraph{Label-aware negative sampling}
In the label-aware variant, the query and positive come from the same training head graph $G_i$. For each node, subgraph, or random-walk triplet, we draw the negative from a head graph $G_j$ in the same training mini-batch only if $y_j \ne y_i$. This excludes same-class cross-graph negatives at all three granularities. If no eligible graph is available, we skip the affected triplet and retain the classification loss.

\paragraph{Multi-view pattern relevance loss}
For each triplet $(q,a,b)$, we first obtain a query-conditioned complement $\mathbf{r}_q=\phi_r(\mathbf{h}_q)$.
Let $\Delta_{qab}=\mathrm{sim}(\mathbf{r}_q,\mathbf{h}_a)-\mathrm{sim}(\mathbf{r}_q,\mathbf{h}_b)$ denote the difference in cosine similarity between the positive and negative examples for query $q$.
For each view, we average the ranking loss over its valid triplets $\mathcal{T}$ in the current mini-batch:
\begin{equation}
\mathcal{L}_{rel}(\mathcal{T}) =
-\frac{1}{|\mathcal{T}|}\sum_{(q,a,b)\in\mathcal{T}}
\log \sigma(\Delta_{qab}).
\label{eq:lrel_set}
\end{equation}
A view with no valid triplets contributes zero to the relevance loss.
Finally, we combine the three granularities as
\begin{equation}
\mathcal{L}_{rel}=
\mu_1 \mathcal{L}_{rel}(\mathcal{T}_{node})
+\mu_2 \mathcal{L}_{rel}(\mathcal{T}_{sub})
+\mu_3 \mathcal{L}_{rel}(\mathcal{T}_{rw}),
\label{eq:lrel_total}
\end{equation}
where $\mu_1,\mu_2,\mu_3$ control the contribution of each view.

\subsection{Graph Complement and Training Objective}
\label{sec:model:objective}
\paragraph{Graph complement}
Given a graph embedding $\mathbf{h}_{G_i}$ produced by the GNN encoder, DMPRP predicts a graph-level complement $\mathbf{r}_{G_i}=\phi_r(\mathbf{h}_{G_i})$ and forms the complemented graph representation
\begin{equation}
\tilde{\mathbf{h}}_{G_i} = \mathbf{h}_{G_i} + \mathbf{r}_{G_i}.
\label{eq:graph_complement}
\end{equation}
We feed $\tilde{\mathbf{h}}_{G_i}$ into a classifier to predict $y_i$.

\paragraph{Task loss}
We optimize the standard cross-entropy loss for graph classification:
\begin{equation}
\mathcal{L}_{task}=
-\sum_{i=1}^{N}\log p\!\left(y_i \mid \tilde{\mathbf{h}}_{G_i}\right).
\label{eq:task_loss}
\end{equation}

\paragraph{Complement dissimilarity}
For a training mini-batch $\mathcal{B}$ containing both head and tail graphs, we average the cosine similarity between each graph embedding and its predicted complement:
\begin{equation}
\mathcal{L}_{dis}=\frac{1}{|\mathcal{B}|}\sum_{G_i\in\mathcal{B}}
\mathrm{sim}(\mathbf{h}_{G_i},\mathbf{r}_{G_i}).
\label{eq:dis_loss}
\end{equation}
Here $\mathrm{sim}(\mathbf{u},\mathbf{v})=\hat{\mathbf{u}}^\top\hat{\mathbf{v}}$, where $\hat{\mathbf{x}}=\mathbf{x}/\max(\lVert\mathbf{x}\rVert_2,10^{-7})$. Minimizing this term discourages positive alignment. A complement $\mathbf{r}=-\mathbf{h}$ can attain the minimum of $-1$ while remaining fully determined by $\mathbf{h}$. The classification and triplet losses provide the task and structural supervision for learning the complement.

\paragraph{Overall objective}
The final training objective of DMPRP is
\begin{equation}
\mathcal{L}=\mathcal{L}_{task}+\mathcal{L}_{rel}+\lambda \mathcal{L}_{dis},
\label{eq:overall_loss}
\end{equation}
where $\lambda$ balances the complement dissimilarity term.
The detailed training procedure is provided in Section~\ref{app:training_procedure}.

\subsection{Detailed Training Procedure}
\label{app:training_procedure}

\begin{algorithm}[H]
\caption{Training procedure of \textbf{DMPRP}}
\label{alg:dmprp}
\begin{algorithmic}[1]
\Require Training dataset $\mathcal{D}=\{(G_i,y_i)\}_{i=1}^N$, head graph set $\mathcal{D}_{head}$, encoder $\phi_g$, relevance predictor $\phi_r$, weights $\mu_1,\mu_2,\mu_3,\lambda$
\While{not converged}
    \State Sample a mini-batch $\mathcal{B}\subset \mathcal{D}$
    \State Compute node embeddings $\{\mathbf{h}_v\}$ and graph embeddings $\{\mathbf{h}_{G}\}$ using $\phi_g$
    \State Sample triplet sets $\mathcal{T}_{node},\mathcal{T}_{sub},\mathcal{T}_{rw}$ from $\mathcal{D}_{head}$
    \State Compute $\mathcal{L}_{rel}$ by Eqs.~\eqref{eq:lrel_set}--\eqref{eq:lrel_total}
    \For{each $G\in \mathcal{B}$}
        \State $\mathbf{r}_{G}\leftarrow \phi_r(\mathbf{h}_{G})$, \quad $\tilde{\mathbf{h}}_{G}\leftarrow \mathbf{h}_{G}+\mathbf{r}_{G}$
    \EndFor
    \State Compute $\mathcal{L}_{task}$ by Eq.~\eqref{eq:task_loss} and $\mathcal{L}_{dis}$ by Eq.~\eqref{eq:dis_loss}
    \State Update parameters of $\phi_g,\phi_r$ by minimizing $\mathcal{L}=\mathcal{L}_{task}+\mathcal{L}_{rel}+\lambda\mathcal{L}_{dis}$
\EndWhile
\end{algorithmic}
\end{algorithm}

Algorithm~\ref{alg:dmprp} summarizes the training procedure.
We first compute node and graph representations using the encoder $\phi_g$.
We then sample multi-granularity triplets from head graphs to compute the multi-view relevance loss $\mathcal{L}_{rel}$.
In parallel, we compute the complemented graph representations for all graphs and optimize the task loss together with the dissimilarity regularizer.

\subsection{Complexity Discussion}
\label{sec:model:complexity}
For a query embedding of dimension $d$, the router maps $d$ inputs to $p$ hidden units and then to $M$ expert scores, at a cost of $O(dp+pM)$. Within each expert, generating the scale and shift matrices costs $O(dp+p^2)$. The dominant operation multiplies the modulated $p\times p$ matrix by a $p\times d$ projection, at a cost of $O(p^2d)$. Pooling the projected rows costs $O(pd)$.

The dense implementation evaluates all $M$ experts and spends $O(Md)$ on their final weighted sum. The total time per query, given its embedding, is therefore $O(dp+pM+Mp^2d+Md)$, dominated by $O(Mp^2d)$. Graph encoding and triplet sampling incur separate costs. Once the complements and embeddings are available, evaluating the two cosine similarities for each of $T$ triplets costs $O(Td)$.

\section{Experiments}
\label{sec:exp}

\subsection{Experimental Setup}
\label{sec:exp:setup}

\subsubsection{Datasets}
\label{sec:exp:datasets}
Following the size-imbalanced LTGC protocol, we evaluate on five widely-used graph classification benchmarks adopted in prior size-imbalanced studies~\cite{qin2025simba}.
They include four bioinformatics datasets and one social network dataset:
\textbf{PTC} (a.k.a.\ PTC-MR), \textbf{FRANK} (a.k.a.\ FRANKENSTEIN), \textbf{D\&D}, \textbf{PROTEINS}, and \textbf{REDDIT} (a.k.a.\ REDDIT-B).
For each dataset, we sort graphs by the number of nodes $|V|$ and define the \emph{head} set as the top 20\% largest graphs.
To avoid splitting graphs with identical sizes at the cutoff, we include all graphs tied with the 20\% threshold, so the head size may be slightly larger than $\lceil 0.2|\mathcal{D}|\rceil$.
The resulting head sizes are:
PTC ($K{=}72$), PROTEINS ($K{=}227$), FRANK ($K{=}922$), D\&D ($K{=}236$), and REDDIT ($K{=}400$).
The remaining graphs are treated as the \emph{tail} set.

\begin{table}[t]
\centering
\small
\caption{Dataset statistics. $K$ denotes the number of head graphs (top 20\% by $|V|$ with ties included).}
\label{tab:datasets}
\resizebox{\columnwidth}{!}{\begin{tabular}{lrrrrrr}
\toprule
Dataset & \#Graphs & Avg.$|V|$ & Avg.$|E|$ & \#Cls & \#Attr & $K$ \\
\midrule
PTC      & 344   & 14.29  & 14.69  & 2 & 18  & 72  \\
FRANK    & 4,337 & 16.90  & 17.88  & 2 & 5   & 922 \\
D\&D     & 1,178 & 284.32 & 715.66 & 2 & 89  & 236 \\
PROTEINS & 1,113 & 39.06  & 72.82  & 2 & 3   & 227 \\
REDDIT   & 2,000 & 429.63 & 497.75 & 2 & 566 & 400 \\
\bottomrule
\end{tabular}}
\end{table}

\subsubsection{Baselines}
\label{sec:exp:baselines}
We compare \textbf{DMPRP} with three categories of baselines widely used for graph classification with size imbalance~\cite{qin2025simba}: 
(i) graph kernels, (ii) standard GNNs with pooling, and (iii) methods designed for imbalance or size shift. 
These baselines cover typical challenges under size imbalance~\cite{oono2019graph}, including size-related shortcuts, distribution shift between head and tail graphs, and weak structural evidence in tail graphs.

\textbf{Graph kernels.}
They measure handcrafted graph similarity without neural training, serving as stable references on small benchmarks but can be sensitive to graph size and costly on large graphs.
(1) \textbf{SP}~\cite{borgwardt2005spkernel} compares graphs using shortest-path length statistics. Computing these statistics becomes more expensive as the numbers of nodes and edges increase.
(2) \textbf{GK}~\cite{shervashidze2009graphlet}: uses counts of small induced subgraphs; motif frequencies are size-dependent and can be noisy for sparse tail graphs.

\textbf{GNNs with pooling.}
They learn representations via message passing and a graph readout. Under size imbalance, pooling may introduce size-related bias and tail graphs may suffer from insufficient neighborhood evidence.
(3) \textbf{GCN}~\cite{kipf2017semi};
(4) \textbf{GIN}~\cite{xu2019gin};
(5) \textbf{DGCNN}~\cite{zhang2018dgcnn} (SortPool);
(6) \textbf{DiffPool}~\cite{ying2018diffpool};
(7) \textbf{ASAPool}~\cite{ranjan2020asap}.

\textbf{Imbalance/size-aware methods.}
They explicitly improve robustness to long-tail regimes or size shift via specialized objectives or regularization.
(8) \textbf{G$^2$GNN}~\cite{wang2022g2gnn};
(9) \textbf{SOLT-GNN}~\cite{liu2022soltgnn};
(10) \textbf{SizeShiftReg}~\cite{buffelli2022sizeshiftreg};
(11) \textbf{SIMBA}~\cite{qin2025simba}.

\subsubsection{Evaluation Metrics}
\label{sec:exp:metrics}
We report \textbf{Accuracy}~\cite{hu2020open} and \textbf{Macro-F1}~\cite{dwivedi2023benchmarking} as the main evaluation metrics, which are standard for graph classification under imbalance~\cite{qin2025simba}.
In addition to overall test performance, we also report results on \textbf{Head} and \textbf{Tail} subsets by evaluating the trained model on the corresponding test graphs.

\subsubsection{Implementation Details}
\label{sec:exp:impl}
We follow the experimental protocol in prior size-imbalanced graph classification work~\cite{qin2025simba}.
For each dataset, we use a \textbf{6:2:2} split for train/validation/test.
All experiments are repeated \textbf{five} times with different random seeds, and we report the mean and standard deviation.
For fair comparison among GNN-based methods, we adopt \textbf{GIN} as the default backbone unless otherwise specified, set the GNN depth to \textbf{3} layers, and use a hidden dimension of \textbf{32} for all methods.
For each baseline, we use the authors' released implementations when available, and tune hyperparameters following their original papers.
For \textbf{DMPRP}, we tune $\{\mu_1,\mu_2,\mu_3,\lambda\}$, the number of experts $M$, and the random-walk length $s$ on the validation set, and report test results under the best validation configuration. All experiments are conducted on a single RTX 5090 GPU with 32GB memory.

\subsection{Overall Performance Comparison}
\label{sec:exp:overall}

\begin{table*}[t]
\centering
\small
\setlength{\tabcolsep}{3.2pt}
\caption{Size-imbalanced graph classification results (Accuracy and Macro-F1, \%) on five datasets.
Bold: best; Underline: runner-up.}
\label{tab:main_results}
\resizebox{\textwidth}{!}{\begin{tabular}{@{} l*{10}{c}}
\toprule
Method
& \multicolumn{2}{c}{PTC}
& \multicolumn{2}{c}{FRANK}
& \multicolumn{2}{c}{D\&D}
& \multicolumn{2}{c}{PROTEINS}
& \multicolumn{2}{c}{REDDIT} \\
\cmidrule(lr){2-3}\cmidrule(lr){4-5}\cmidrule(lr){6-7}\cmidrule(lr){8-9}\cmidrule(lr){10-11}
& Acc. & M-F1 & Acc. & M-F1 & Acc. & M-F1 & Acc. & M-F1 & Acc. & M-F1 \\
\midrule
SP
& 56.00\stdpm{1.40} & 47.29\stdpm{3.56}
& 59.10\stdpm{1.85} & 53.01\stdpm{1.93}
& 75.30\stdpm{1.61} & 72.41\stdpm{1.78}
& 73.66\stdpm{1.86} & 71.13\stdpm{1.99}
& 75.05\stdpm{1.76} & 74.78\stdpm{1.63} \\
GK
& 58.16\stdpm{4.90} & 46.75\stdpm{6.02}
& 57.83\stdpm{1.86} & 50.36\stdpm{2.60}
& 72.57\stdpm{1.13} & 69.27\stdpm{4.78}
& 71.25\stdpm{1.46} & 66.32\stdpm{2.42}
& 71.75\stdpm{2.17} & 70.25\stdpm{2.70} \\
\midrule
GCN
& 54.29\stdpm{4.69} & 46.69\stdpm{8.25}
& 66.61\stdpm{1.84} & 65.05\stdpm{2.07}
& 75.02\stdpm{4.05} & 72.28\stdpm{3.46}
& 72.74\stdpm{1.51} & 66.38\stdpm{2.48}
& 79.60\stdpm{2.32} & 78.07\stdpm{3.44} \\
GIN
& 57.43\stdpm{8.05} & 43.15\stdpm{6.34}
& 70.63\stdpm{1.37} & 65.80\stdpm{1.34}
& 69.70\stdpm{5.04} & 55.95\stdpm{6.29}
& 73.63\stdpm{3.78} & 61.97\stdpm{11.53}
& 66.35\stdpm{9.16} & 62.59\stdpm{10.24} \\
DGCNN
& 57.43\stdpm{3.88} & 41.99\stdpm{7.47}
& 66.65\stdpm{1.55} & 66.24\stdpm{1.65}
& 63.29\stdpm{3.75} & 56.11\stdpm{5.30}
& 70.58\stdpm{2.84} & 66.07\stdpm{2.33}
& 70.45\stdpm{2.86} & 69.95\stdpm{6.61} \\
DiffPool
& 57.29\stdpm{5.26} & 47.07\stdpm{6.68}
& 66.78\stdpm{1.98} & 66.99\stdpm{2.03}
& 66.94\stdpm{4.50} & 66.32\stdpm{4.19}
& 70.69\stdpm{2.58} & 65.69\stdpm{1.66}
& 74.95\stdpm{3.05} & 74.52\stdpm{6.11} \\
ASAPool
& 58.00\stdpm{5.24} & 46.53\stdpm{8.33}
& 66.10\stdpm{1.07} & 65.07\stdpm{2.23}
& 71.31\stdpm{4.04} & 68.21\stdpm{2.55}
& 72.47\stdpm{1.67} & 62.36\stdpm{3.02}
& 75.75\stdpm{5.77} & 71.04\stdpm{5.64} \\
\midrule
G2GNN
& 56.86\stdpm{7.79} & 52.56\stdpm{9.81}
& 71.54\stdpm{4.78} & 64.08\stdpm{2.11}
& 71.22\stdpm{3.69} & 69.27\stdpm{4.78}
& 65.54\stdpm{1.92} & 64.35\stdpm{2.02}
& 78.10\stdpm{1.23} & 80.97\stdpm{1.53} \\
SOLT-GNN
& 56.47\stdpm{3.83} & 53.85\stdpm{6.58}
& 71.34\stdpm{1.06} & \underline{67.96\stdpm{2.16}}
& 75.20\stdpm{1.76} & 72.07\stdpm{3.07}
& \underline{74.23\stdpm{2.24}} & 71.67\stdpm{2.23}
& 77.20\stdpm{5.09} & 80.43\stdpm{4.89} \\
SizeShiftReg
& 57.07\stdpm{0.95} & 53.34\stdpm{5.56}
& 70.46\stdpm{1.39} & 66.17\stdpm{2.67}
& 70.42\stdpm{3.46} & 72.05\stdpm{2.67}
& 72.39\stdpm{2.57} & 69.13\stdpm{2.89}
& 79.35\stdpm{6.97} & 79.37\stdpm{2.48} \\
SIMBA
& \underline{59.43\stdpm{5.97}} & \underline{55.12\stdpm{6.20}}
& \underline{72.69\stdpm{1.56}} & 67.04\stdpm{1.17}
& \underline{75.95\stdpm{5.38}} & \underline{73.21\stdpm{3.09}}
& 74.11\stdpm{4.23} & \underline{72.15\stdpm{1.77}}
& \underline{82.70\stdpm{4.20}} & \underline{81.28\stdpm{4.21}} \\
\textbf{DMPRP}
& \textbf{63.88\stdpm{6.65}} & \textbf{61.55\stdpm{5.37}}
& \textbf{73.59\stdpm{0.92}} & \textbf{72.83\stdpm{1.03}}
& \textbf{77.53\stdpm{0.87}} & \textbf{77.13\stdpm{0.94}}
& \textbf{75.14\stdpm{0.92}} & \textbf{73.83\stdpm{1.42}}
& \textbf{85.05\stdpm{1.51}} & \textbf{84.94\stdpm{1.54}} \\
\bottomrule
\end{tabular}}
\end{table*}

\begin{table*}[t]
\centering
\normalsize
\setlength{\tabcolsep}{4pt}
\caption{Accuracy (\%)$\pm$std on the overall test set (\textbf{Test}), topology-rich head subset (\textbf{Head}), and topology-scarce tail subset (\textbf{Tail}) under the LTGC protocol.
Bold: best.}
\label{tab:head_tail_acc}

\begingroup
\sisetup{
  detect-all,
  table-number-alignment = center
}
\newcommand{\sbest}[1]{{\bfseries #1}}

\resizebox{\textwidth}{!}{\begin{tabular}{@{} l|
*{3}{M@{\spm}U}|
*{3}{M@{\spm}U}|
*{3}{M@{\spm}U}}
\toprule
\multirow{2}{*}{Method}
& \multicolumn{6}{c|}{PTC}
& \multicolumn{6}{c|}{FRANK}
& \multicolumn{6}{c}{D\&D} \\
& \multicolumn{2}{c}{Test} & \multicolumn{2}{c}{Head} & \multicolumn{2}{c|}{Tail}
& \multicolumn{2}{c}{Test} & \multicolumn{2}{c}{Head} & \multicolumn{2}{c|}{Tail}
& \multicolumn{2}{c}{Test} & \multicolumn{2}{c}{Head} & \multicolumn{2}{c}{Tail} \\
\midrule
GIN
& 57.43 & 8.05 & 57.68 & 11.30 & 56.39 & 6.96
& 70.63 & 1.37 & 72.32 & 5.81 & 70.31 & 1.32
& 69.70 & 5.04 & 84.27 & 4.47 & 66.31 & 5.38 \\
SOLT-GNN
& 56.47 & 3.83 & 61.05 & 10.90 & 54.69 & 5.66
& 71.34 & 1.06 & 73.52 & 5.23 & 70.34 & 1.04
& 75.20 & 1.76 & 79.26 & 4.87 & 73.94 & 1.98 \\
SizeShiftReg
& 57.07 & 6.95 & 62.53 & 7.05 & 56.96 & 7.30
& 70.46 & 1.39 & 70.40 & 1.35 & 70.70 & 2.60
& 70.42 & 3.46 & 83.02 & 6.04 & 67.27 & 3.44 \\
SIMBA
& 59.43 & 5.97 & 67.77 & 6.20 & 57.29 & 7.44
& 72.69 & 1.56 & \sbest{78.57} & 1.37 & 71.11 & 2.11
& 75.95 & 3.38 & \sbest{88.27} & 2.93 & 72.73 & 4.66 \\
DMPRP
& \sbest{63.88} & 6.65 & \sbest{70.63} & 6.61 & \sbest{62.50} & 6.35
& \sbest{73.59} & 0.92 & 74.40 & 0.93 & \sbest{73.37} & 1.31
& \sbest{77.53} & 0.87 & 85.07 & 0.96 & \sbest{75.64} & 0.95 \\
\bottomrule
\end{tabular}}

\endgroup
\end{table*}

Table~\ref{tab:main_results} compares the methods on five size-imbalanced benchmarks. The strongest baseline varies with the dataset and metric. SIMBA leads the baselines on both metrics on PTC, D\&D, and REDDIT. On FRANK, SIMBA has the highest baseline accuracy, while SOLT-GNN has the highest baseline Macro-F1. These size-aware models also outperform the graph kernels and the standard GIN encoder on both metrics on FRANK.

DMPRP achieves the highest mean test accuracy and Macro-F1 on all five datasets. Its clearest advantage over the strongest baseline is on PTC, where it gains 6.43 percentage points in Macro-F1 over SIMBA. On FRANK, the gain in accuracy is modest, but the margin in Macro-F1 over SOLT-GNN is larger. On PROTEINS, the gain in accuracy over SOLT-GNN is less than one percentage point. The ablations in Section~\ref{sec:exp:ablation} examine how the expert module and the different sources of triplet supervision contribute to these results.

\subsection{Long-tailed Graph Classification Protocol}
\label{sec:exp:ltgc_protocol}
Following the size-oriented LTGC setting~\cite{liu2022soltgnn}, our evaluation mainly focuses on improving performance on topology-scarce \emph{tail} graphs.
Accordingly, we report both the overall test performance (\textbf{Test}) and the tail-subset performance (\textbf{Tail}) across datasets, where graphs are ranked by size and split into head (top 20\%) and tail (bottom 80\%) subsets.
In our implementation, \textbf{DMPRP} activates the dynamic memory-based pattern complement only for tail graphs, where transferable pattern evidence is most needed; for topology-rich head graphs, we directly employ the base GNN encoder representation for prediction since their abundant structures are already sufficiently informative.
Formally, we use $\hat{y}=f(\tilde{\mathbf{h}}_{G})$ for tail graphs and $\hat{y}=f(\mathbf{h}_{G})$ for head graphs, where $\tilde{\mathbf{h}}_{G}$ is the complemented representation and $\mathbf{h}_{G}$ is the base encoder output.

Table~\ref{tab:head_tail_acc} shows that on PTC, DMPRP improves mean accuracy over SIMBA by 5.21 percentage points on tail graphs, nearly twice the gain on head graphs. This narrows the gap between the two groups, although both methods still classify head graphs more accurately.

On FRANK and D\&D, DMPRP achieves the highest mean tail accuracy among the methods compared, while SIMBA leads on head graphs. On these two datasets, DMPRP's advantage in overall accuracy thus coexists with lower accuracy than SIMBA on the larger graphs.

\subsection{Ablation Study}
\label{sec:exp:ablation}

We conduct ablation studies to assess the contribution of each key component in \textbf{DMPRP}.
All variants share the same encoder backbone, data splits, and optimization settings; we only remove one component at a time while keeping all other modules unchanged.
Specifically, we evaluate: (i) \textbf{w/o RW}, removing the neighborhood-level random-walk triplets and the associated loss; (ii) \textbf{w/o Node}, removing node-level triplet supervision; (iii) \textbf{w/o Subgraph}, removing subgraph-level triplet supervision; and (iv) \textbf{w/o Experts}, disabling the KDEX-style dynamic expert memory and replacing it with a non-retrieval predictor.

As shown in Table~\ref{tab:ablation_5datasets}, each component contributes positively, and the full \textbf{DMPRP} model achieves the best and most stable performance across datasets and splits.

\begin{table*}[t]
\centering
\small
\setlength{\tabcolsep}{3.2pt}
\caption{Ablation results of \textbf{DMPRP} on five datasets. We report Accuracy and Macro-F1 (\%) on the overall test set (\textbf{Test}) and the topology-scarce tail subset (\textbf{Tail}).}
\label{tab:ablation_5datasets}

\resizebox{\textwidth}{!}{\begin{tabular}{@{} l l
*{5}{ M@{\spm}U M@{\spm}U } }
\toprule
Variant & Split
& \multicolumn{4}{c}{PTC}
& \multicolumn{4}{c}{FRANK}
& \multicolumn{4}{c}{D\&D}
& \multicolumn{4}{c}{PROTEINS}
& \multicolumn{4}{c}{REDDIT} \\
\cmidrule(lr){3-6}\cmidrule(lr){7-10}\cmidrule(lr){11-14}\cmidrule(lr){15-18}\cmidrule(lr){19-22}
&
& \multicolumn{2}{c}{Acc.} & \multicolumn{2}{c}{M-F1}
& \multicolumn{2}{c}{Acc.} & \multicolumn{2}{c}{M-F1}
& \multicolumn{2}{c}{Acc.} & \multicolumn{2}{c}{M-F1}
& \multicolumn{2}{c}{Acc.} & \multicolumn{2}{c}{M-F1}
& \multicolumn{2}{c}{Acc.} & \multicolumn{2}{c}{M-F1} \\
\midrule

Full & Test
& 63.88 & 6.65 & 61.55 & 5.37
& 73.59 & 0.92 & 72.83 & 1.03
& 77.53 & 0.87 & 77.13 & 0.94
& 75.14 & 0.92 & 73.83 & 1.42
& 85.05 & 1.51 & 84.94 & 1.54 \\
     & Tail
& 62.50 & 6.35 & 61.82 & 7.00
& 73.37 & 1.31 & 72.70 & 1.47
& 75.64 & 0.95 & 75.51 & 1.02
& 73.56 & 0.84 & 73.24 & 1.40
& 87.68 & 1.13 & 87.18 & 1.25 \\

\midrule
w/o RW & Test
& 56.47 & 7.40 & 55.29 & 7.49
& 72.83 & 1.49 & 72.37 & 1.42
& 73.62 & 2.17 & 73.19 & 2.28
& 73.87 & 1.46 & 72.30 & 1.85
& 82.15 & 1.83 & 82.04 & 1.81 \\
       & Tail
& 60.00 & 9.85 & 59.74 & 10.01
& 72.62 & 1.51 & 72.27 & 1.45
& 71.17 & 2.95 & 70.86 & 3.00
& 73.03 & 1.64 & 72.53 & 1.78
& 84.23 & 1.30 & 83.72 & 1.38 \\

w/o Node & Test
& 55.00 & 8.68 & 53.91 & 7.78
& 58.71 & 6.69 & 50.79 & 6.64
& 56.51 & 1.66 & 47.35 & 9.23
& 56.13 & 2.15 & 51.31 & 7.27
& 82.10 & 1.10 & 82.05 & 1.11 \\
        & Tail
& 56.33 & 8.49 & 56.08 & 8.24
& 55.14 & 8.47 & 42.25 & 11.20
& 50.53 & 1.50 & 33.92 & 0.81
& 50.63 & 1.25 & 38.35 & 7.52
& 83.69 & 1.18 & 83.37 & 1.29 \\

w/o Subgraph & Test
& 59.41 & 8.23 & 58.69 & 7.54
& 72.30 & 0.58 & 71.48 & 0.45
& 76.60 & 2.06 & 76.09 & 2.06
& 73.42 & 1.80 & 71.91 & 2.14
& 82.65 & 1.60 & 82.43 & 1.57 \\
           & Tail
& 59.59 & 8.83 & 58.79 & 9.07
& 71.84 & 0.63 & 71.08 & 0.53
& 74.79 & 1.83 & 74.74 & 1.84
& 71.54 & 2.87 & 70.96 & 3.26
& 84.46 & 1.46 & 83.68 & 1.79 \\

w/o Experts & Test
& 61.76 & 7.57 & 59.34 & 7.56
& 72.50 & 1.64 & 71.82 & 1.90
& 72.51 & 4.20 & 70.89 & 6.09
& 73.78 & 1.33 & 72.49 & 1.32
& 79.15 & 3.29 & 78.95 & 3.36 \\
          & Tail
& 59.59 & 8.34 & 58.70 & 8.57
& 72.19 & 2.04 & 71.51 & 2.32
& 70.74 & 4.68 & 70.04 & 6.08
& 72.46 & 1.53 & 72.05 & 1.53
& 80.36 & 5.01 & 79.67 & 4.83 \\

\bottomrule
\end{tabular}}
\end{table*}
Removing \textbf{random-walk triplets} lowers mean tail accuracy on all five datasets. D\&D shows the largest drop, at 4.47 percentage points. PTC and REDDIT also lose tail accuracy and Macro-F1 even though their models still receive node and subgraph supervision. These comparisons suggest that the neighborhoods sampled by random walks contribute useful information beyond the other two views.

Removing \textbf{node-level triplets} has a particularly pronounced effect on FRANK and D\&D. On D\&D, test Macro-F1 falls from 77.13\% to 47.35\%, the largest decline among the four ablations on this dataset. FRANK shows a similar dependence on node supervision. On REDDIT the decline is smaller, but it appears on both the full test set and the tail subset.

The contribution of \textbf{subgraph-level triplets} is smaller than that of node triplets on most datasets. On PROTEINS, omitting subgraph supervision lowers both accuracy and Macro-F1 on the full test set, and the decline is also visible among tail graphs. REDDIT follows the same pattern. In both cases the full model benefits from learning which subgraphs co-occur within a graph.

Replacing the \textbf{KDEX-style dynamic expert memory} has its clearest effect on REDDIT, where mean tail accuracy drops by 7.32 percentage points. This exceeds the drop from removing any single triplet view on that dataset. The expert ablation also lowers both tail metrics on D\&D, although removing node supervision has a larger effect there.

\subsection{Hyperparameter Analysis}
\label{sec:exp:hyper}
We analyze the sensitivity of DMPRP to three key hyperparameters in the main text:
(i) the number of experts $M$ in the expert-based dynamic memory,
(ii) the random-walk length $s$ for neighborhood-level triplet mining, and
(iii) the weight $\lambda$ of the complement dissimilarity regularizer.

Additional analysis of the multi-view loss coefficients and the numbers of sampled triplets is provided in Section~\ref{app:additional_hyper}.

\begin{figure*}[t]
  \centering
  \begin{tabular}{@{}ccc@{}}
    \small Expert count $M$ & \small Random-walk length $s$ & \small Dissimilarity weight $\lambda$ \\
    \includegraphics[trim={-3.792bp -3.571bp -4.834bp -3.912bp},width=.316\textwidth,height=.200\textwidth]{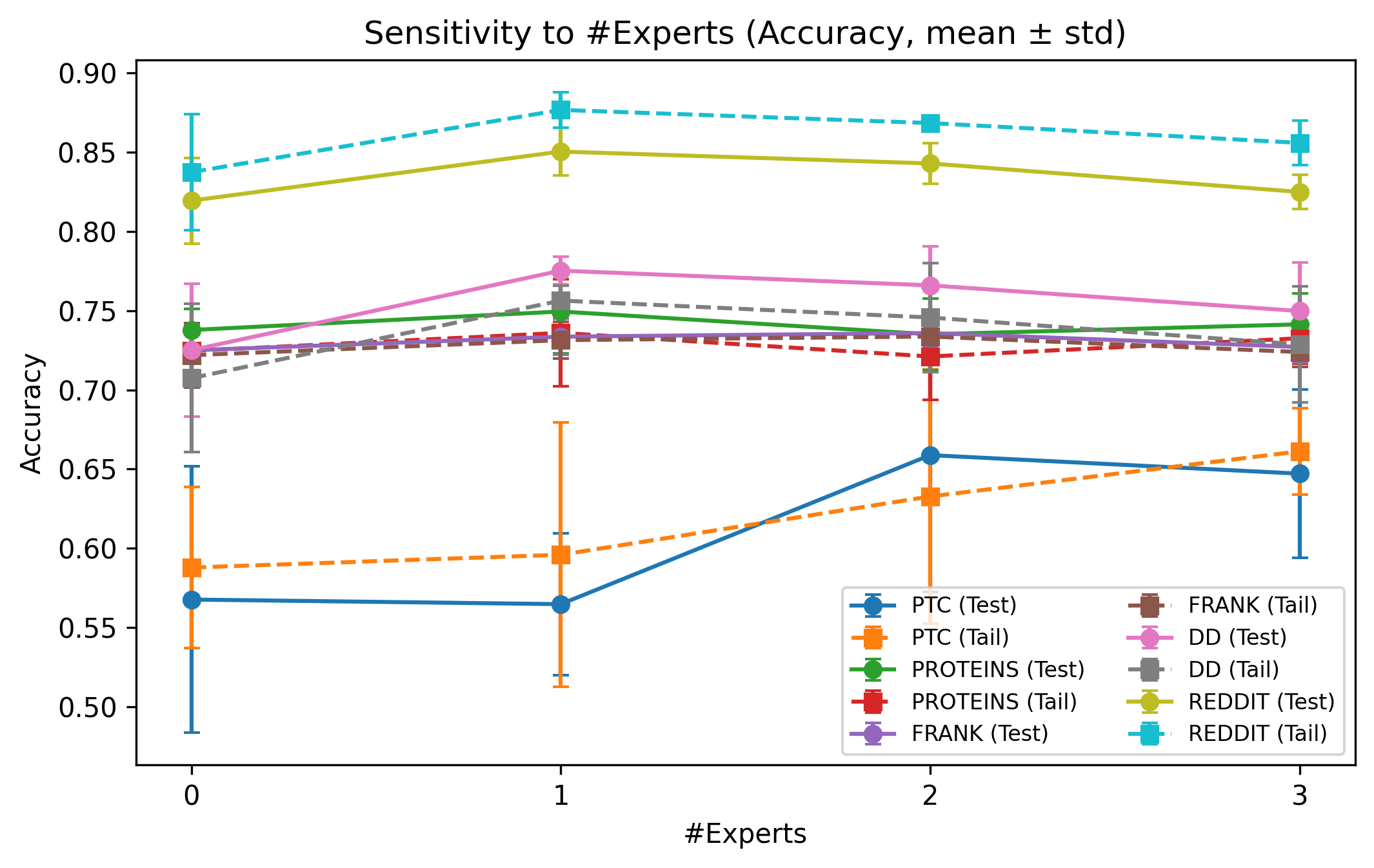} &
    \includegraphics[trim={-17.309bp -8.602bp -5.838bp -21.816bp},width=.316\textwidth,height=.200\textwidth]{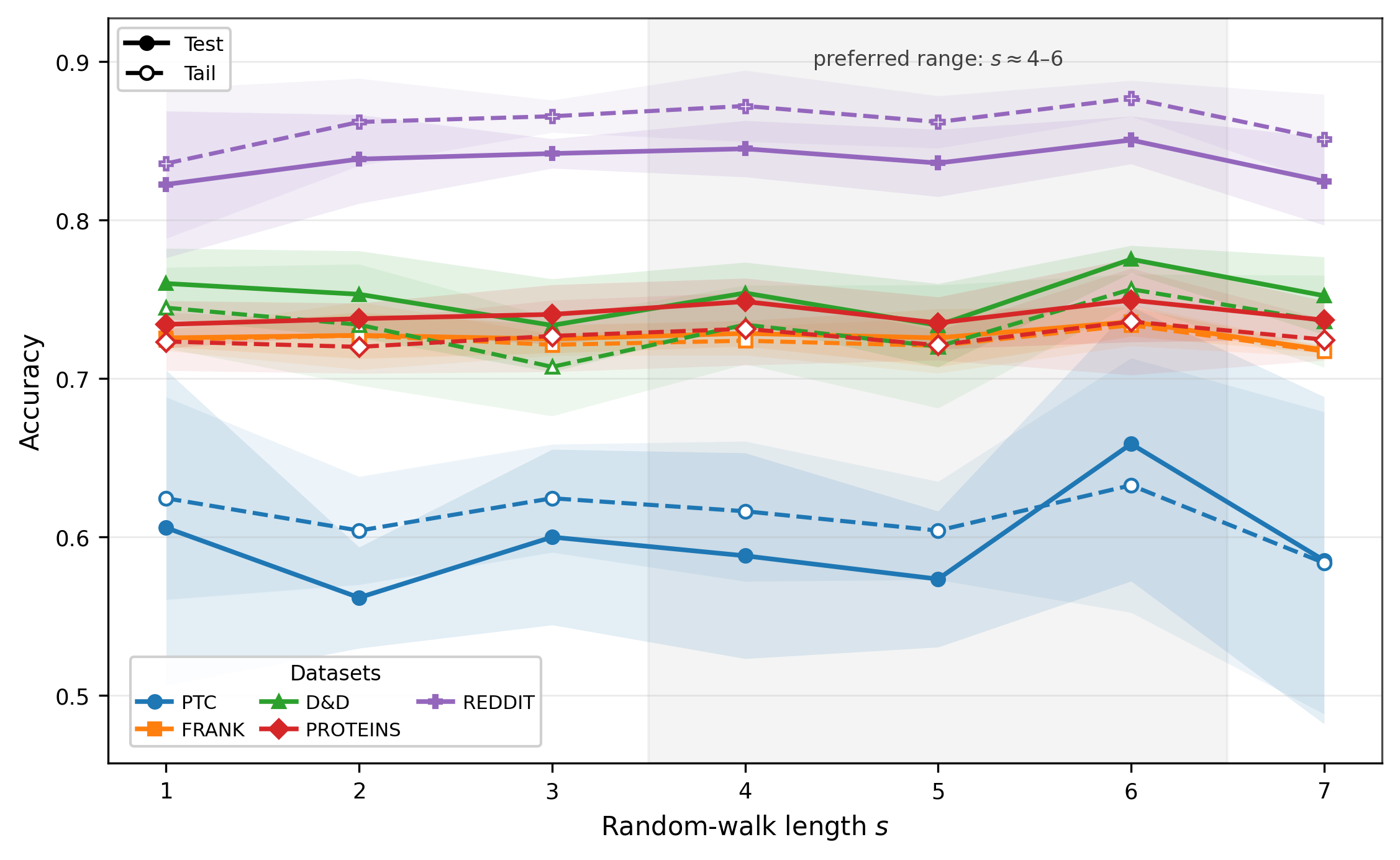} &
    \includegraphics[trim={-19.048bp -8.889bp -7.564bp -8.051bp},width=.316\textwidth,height=.200\textwidth]{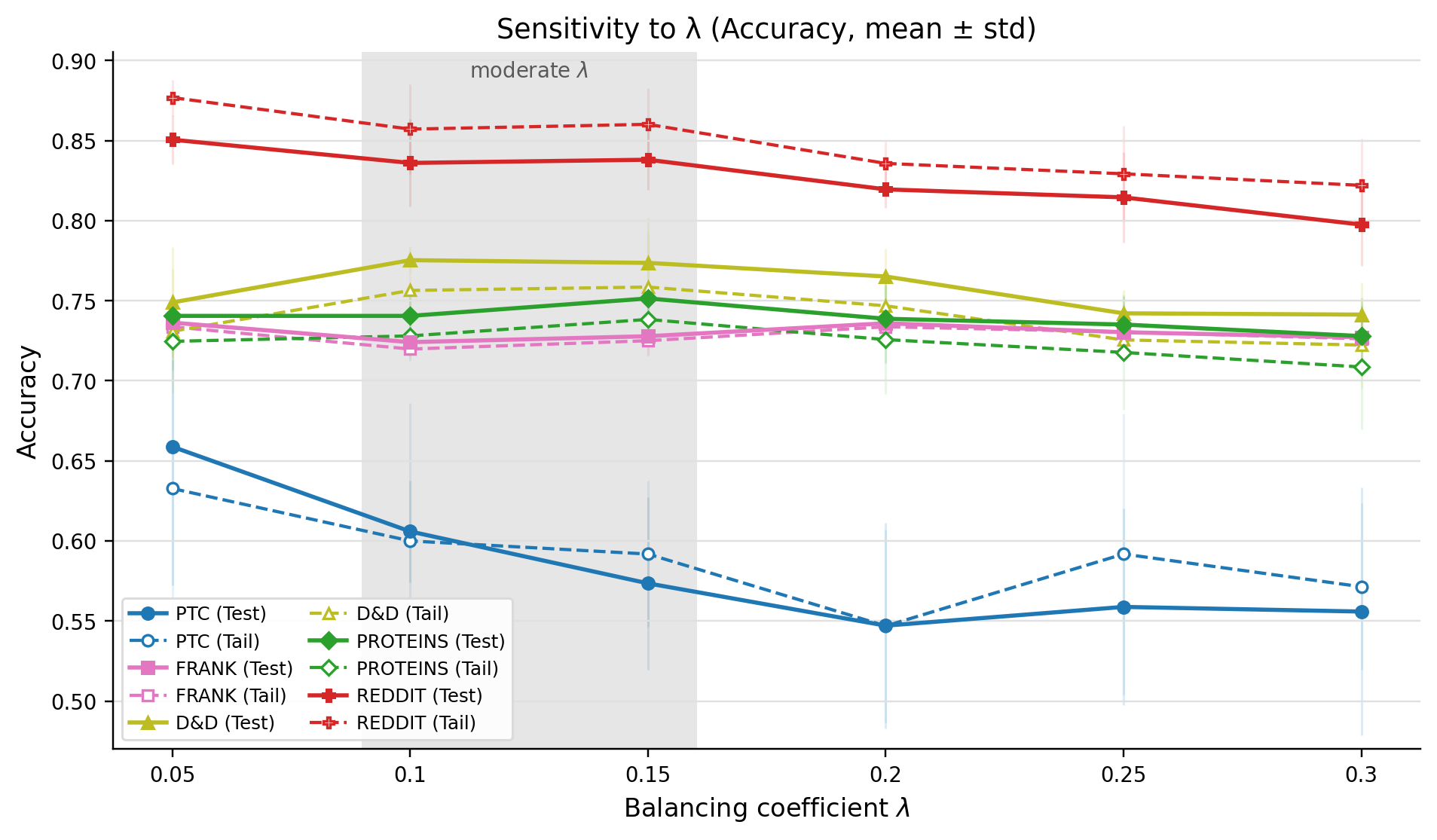} \\
    \includegraphics[trim={-3.792bp -3.571bp -4.834bp -3.912bp},width=.316\textwidth,height=.200\textwidth]{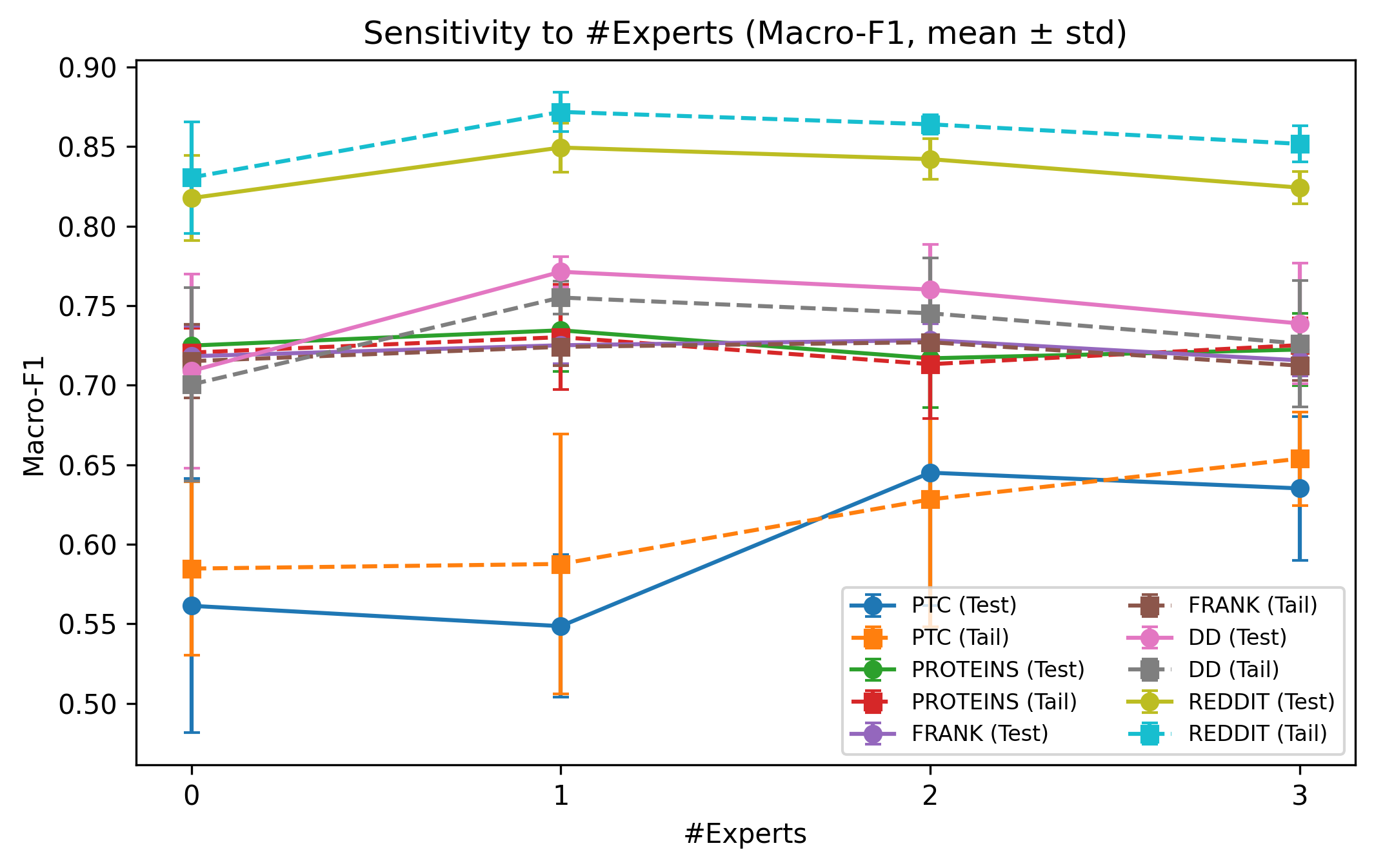} &
    \includegraphics[trim={-17.309bp -8.602bp -5.838bp -21.816bp},width=.316\textwidth,height=.200\textwidth]{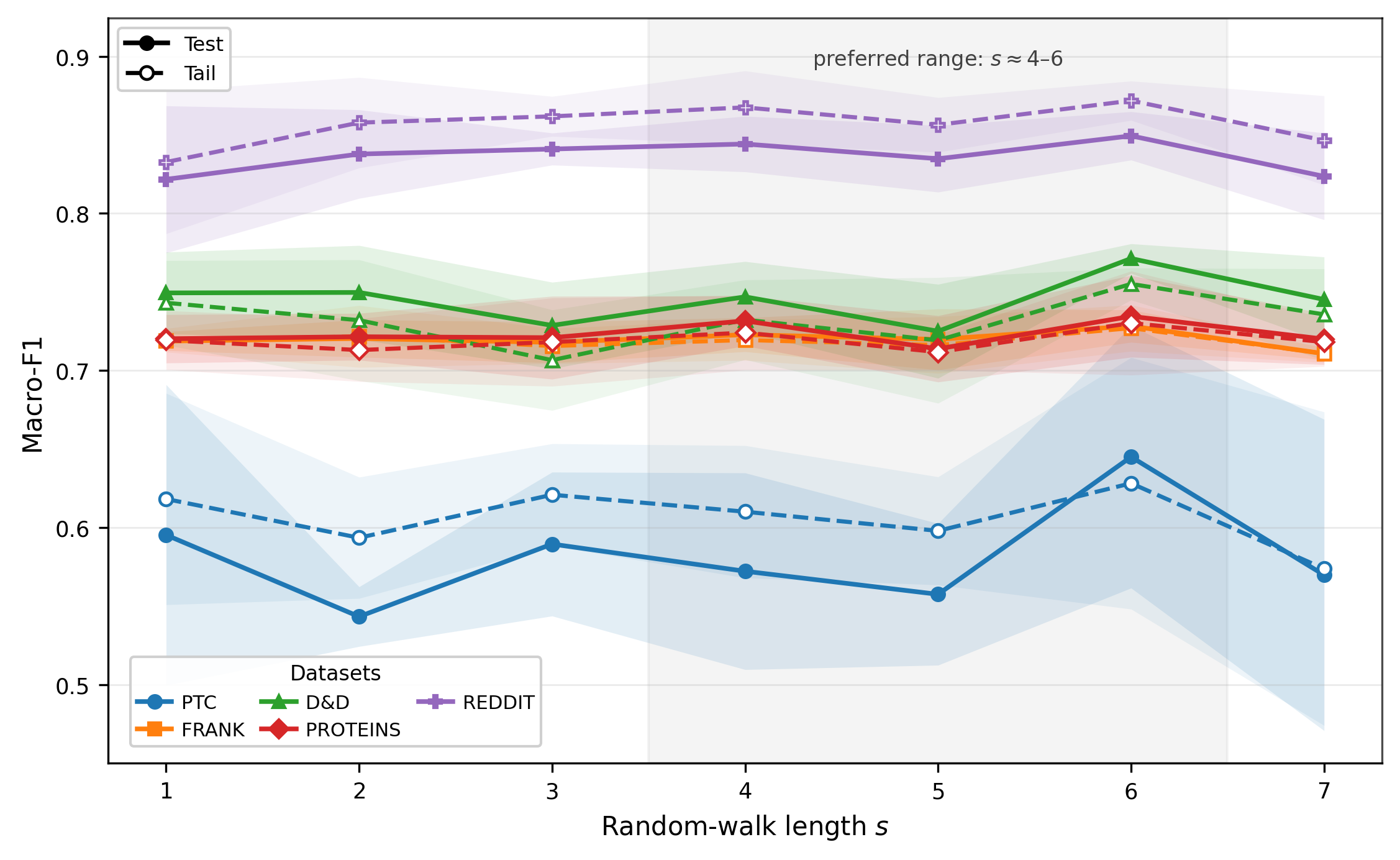} &
    \includegraphics[trim={-19.048bp -8.889bp -7.564bp -8.051bp},width=.316\textwidth,height=.200\textwidth]{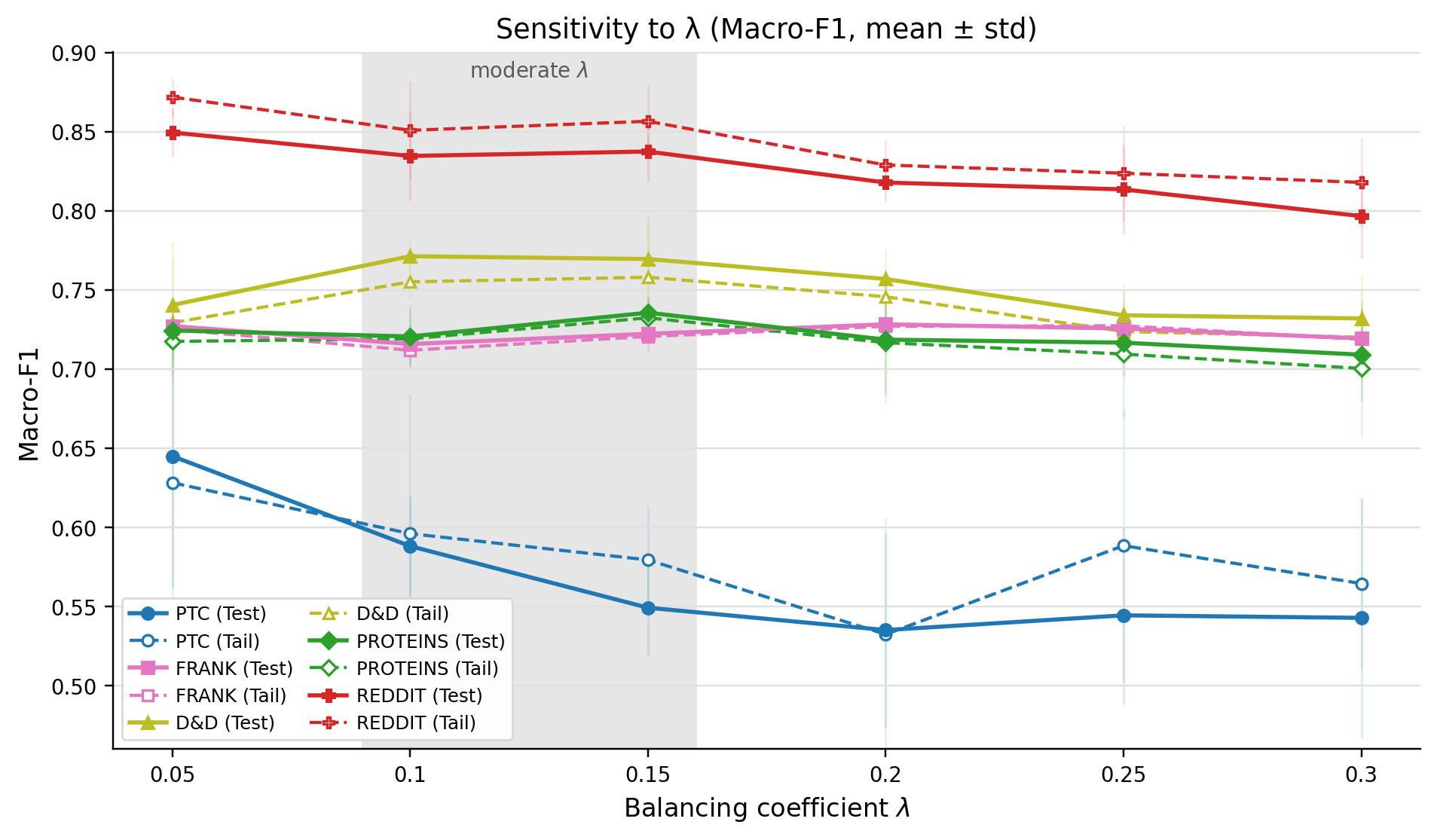}
  \end{tabular}
  \caption{Hyperparameter sensitivity on five datasets. The top row shows Accuracy and the bottom row shows Macro-F1. Solid and dashed lines show means on the full test set and the tail subset, respectively. Error bars in the left column and colored bands in the middle column indicate one standard deviation. Gray shading highlights walk lengths $s=4$--$6$ and dissimilarity weights $\lambda=0.10$--$0.15$.}
  \label{fig:hyper_joint}
\end{figure*}

\textbf{Number of experts.}
We sweep $M \in \{0,1,2,3\}$ and summarize the results in the left column of Fig.~\ref{fig:hyper_joint}.
Overall, introducing experts consistently improves both Accuracy and Macro-F1 on most datasets, with particularly clear gains on topology-scarce tail graphs.
Meanwhile, the improvement tends to \emph{saturate} when increasing $M$ beyond a moderate value: $M=1$ or $M=2$ typically provides the best trade-off between performance and stability, while using more experts (e.g., $M=3$) yields marginal benefits or even slight fluctuations.
This suggests that an overly large expert pool may introduce noisy retrieval/selection and weaken the robustness of complement prediction.

\textbf{Random-walk length.}
We vary the random-walk length $s \in \{1,2,3,4,5,6,7\}$ for neighborhood-level triplet mining (Fig.~\ref{fig:hyper_joint}, middle column).
Overall, DMPRP is relatively robust to $s$, but a \emph{moderate} walk length consistently yields stronger and more stable performance.
In particular, $s$ around $4$--$6$ achieves the best or near-best Accuracy and Macro-F1 on most datasets, with a clear tendency that $s{=}6$ improves the tail subset across benchmarks.
When $s$ is too small (e.g., $s\le 2$), the mined neighborhoods contain limited context and the benefit of relevance learning is reduced; when $s$ becomes too large (e.g., $s{=}7$), the gain saturates and may slightly drop due to introducing noisier/less relevant nodes.
Based on these observations, we adopt a moderate setting (default $s{=}6$) in all experiments.

\textbf{Complement dissimilarity weight.}
We vary $\lambda \in \{0.05, 0.10, 0.15, 0.20, 0.25, 0.30\}$ in Eq.~\eqref{eq:overall_loss}. Increasing $\lambda$ gives more weight to the penalty on cosine alignment between the graph embedding and its complement. The random-walk triplet loss has its own weight, $\mu_3$, in Eq.~\eqref{eq:lrel_total}. The right column of Fig.~\ref{fig:hyper_joint} shows that the response varies by dataset. For example, PTC attains its highest test Accuracy and Macro-F1 at $\lambda=0.05$, while several other datasets show little change between $0.10$ and $0.15$.

\subsection{Additional Hyperparameter Analysis}
\label{app:additional_hyper}

We examine sensitivity to the multi-view relevance loss coefficients and the numbers of sampled triplets. 
Unless otherwise specified, we vary one hyperparameter at a time while keeping the others fixed.

\subsubsection{Multi-view Loss Coefficients}
\label{app:hyper_mu}
\begin{figure*}[t]
  \centering
  \includegraphics[width=\textwidth]{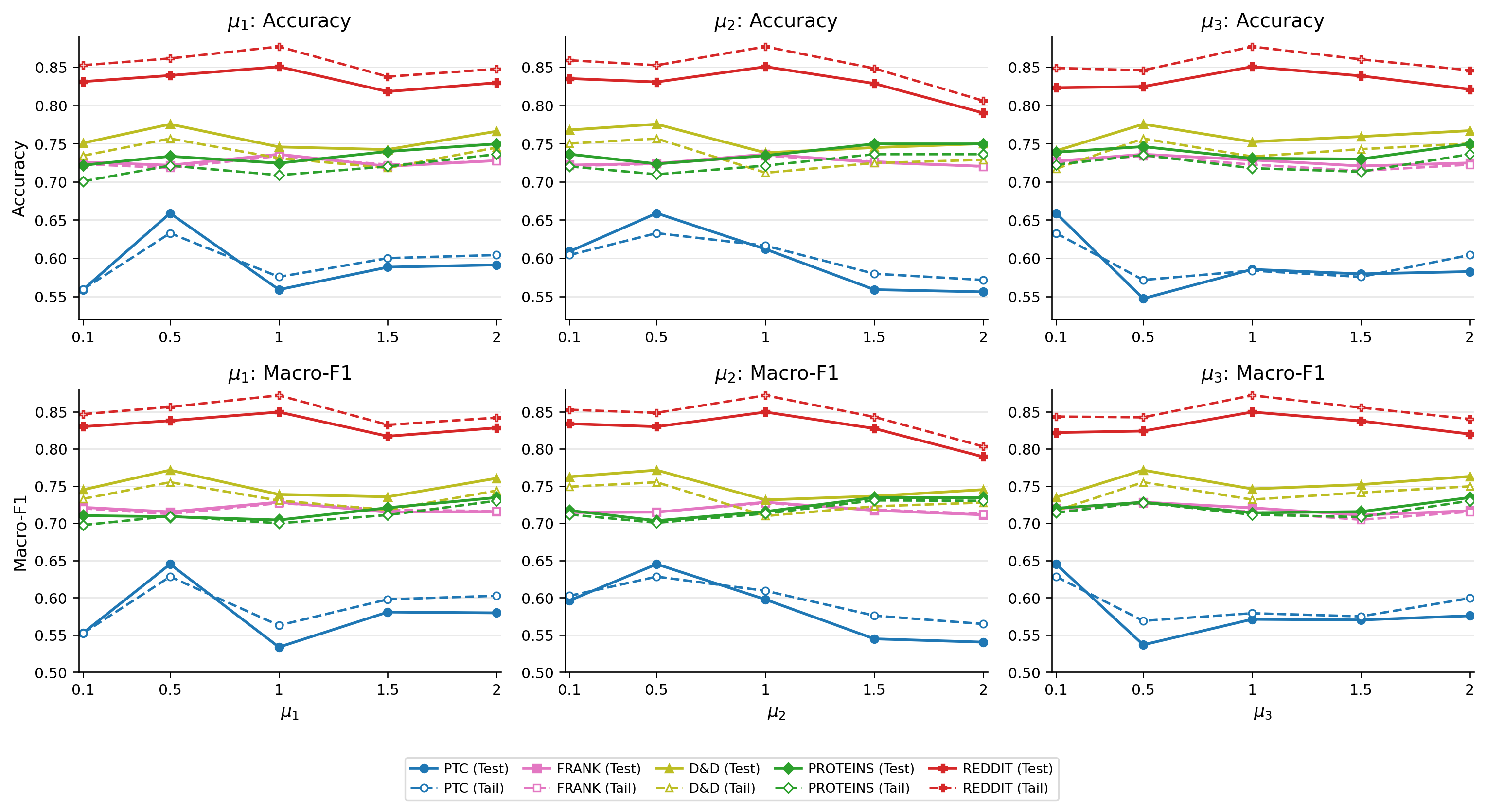}
  \caption{
  Sensitivity to the multi-view loss coefficients $\mu_1$, $\mu_2$, and $\mu_3$ on five datasets.
  The top row shows Accuracy and the bottom row shows Macro-F1.
  From left to right, the three columns correspond to $\mu_1$, $\mu_2$, and $\mu_3$, which control the node-level, subgraph-level, and neighborhood-level random-walk relevance losses in Eq.~(10), respectively.
  Solid lines denote Test and dashed lines denote Tail.
  }
  \label{fig:hyper_mu}
\end{figure*}

To examine the role of each structural view, we vary $\mu_1$, $\mu_2$, and $\mu_3$ in Eq.~(10). These coefficients correspond to the node-level relevance loss, the subgraph-level relevance loss, and the random-walk relevance loss, respectively.
Fig.~\ref{fig:hyper_mu} reports Test and Tail Accuracy/Macro-F1 on five datasets when one coefficient is varied over $\{0.1,0.5,1,1.5,2\}$.

Overall, DMPRP remains reasonably robust under moderate coefficient settings, while overly small or overly large weights may weaken the balance among different structural views.
Although the best value is dataset-dependent, Fig.~\ref{fig:hyper_mu} shows a consistent trend that properly balancing the three relevance views is important for achieving strong and stable performance on both Test and Tail.

For $\mu_1$, which weights the node-level relevance loss, moderate values generally yield stronger results.
PTC and D\&D achieve their best overall performance around $\mu_1=0.5$, whereas FRANK and REDDIT perform best near $\mu_1=1$.
PROTEINS is relatively stable and tends to benefit from a somewhat larger $\mu_1$.
These observations indicate that node-level supervision is essential, while its optimal strength depends on the discriminability of local structural patterns in different datasets.

For $\mu_2$, which controls the subgraph-level relevance loss, we observe a similar moderate-range preference.
PTC and D\&D favor $\mu_2=0.5$, FRANK and REDDIT are strongest around $\mu_2=1$, whereas PROTEINS reaches its best performance at relatively larger values.
At the same time, over-emphasizing subgraph-level supervision can reduce robustness on some datasets; for example, REDDIT drops noticeably when $\mu_2$ becomes too large.

For $\mu_3$, which weights the neighborhood-level random-walk relevance loss, the preferred value is more dataset-dependent.
PTC tends to prefer a relatively small $\mu_3$, FRANK is most stable around $\mu_3=0.5$, REDDIT favors $\mu_3=1$, and PROTEINS benefits from a larger value.
This suggests that the contribution of neighborhood-level supervision varies with graph topology and the amount of transferable local context available in each dataset.

Overall, the results in Fig.~\ref{fig:hyper_mu} confirm that the three views in the multi-granularity relevance loss are complementary.
In practice, moderate coefficients generally provide a good trade-off between effectiveness and stability, while validation-based tuning of $\mu_1$, $\mu_2$, and $\mu_3$ remains beneficial for obtaining the best final configuration.

\begin{figure*}[t]
  \centering
  \includegraphics[width=\textwidth]{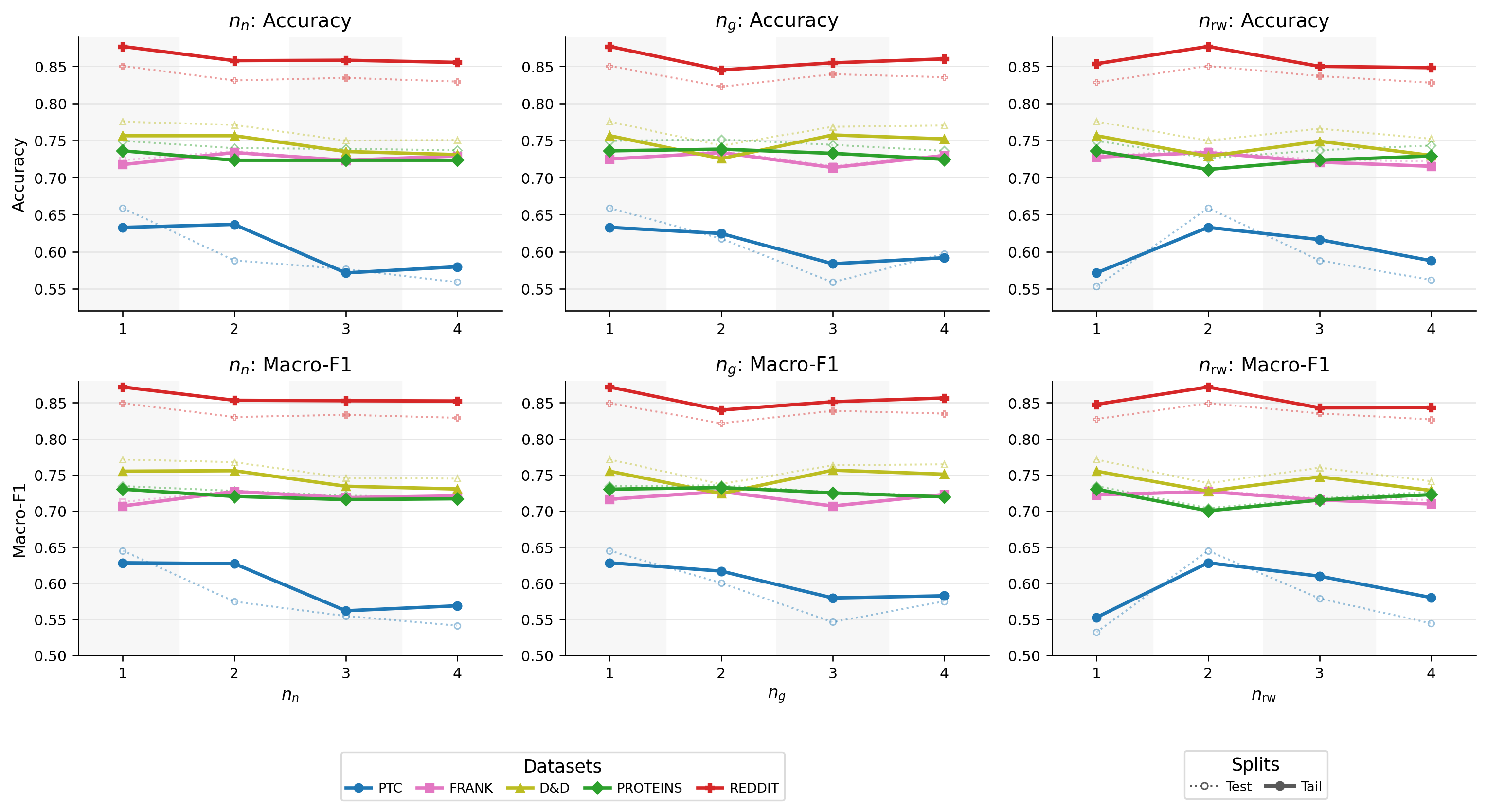}
  \caption{
  Sensitivity to the numbers of sampled triplets on five datasets.
  The top row shows Accuracy and the bottom row shows Macro-F1.
  From left to right, the three columns correspond to the numbers of node-level triplets $n_n$, subgraph-level triplets $n_g$, and random-walk triplets $n_{\mathrm{rw}}$, respectively.
  Dotted lines with open markers show results on the full test set. Solid lines with filled markers show results on the tail subset.
  }
  \label{fig:hyper_triplet_counts}
\end{figure*}

\subsubsection{Numbers of Sampled Triplets}
\label{app:hyper_triplet_counts}

We further analyze the sensitivity of DMPRP to the numbers of sampled node-level triplets $n_n$, subgraph-level triplets $n_g$, and random-walk triplets $n_{\mathrm{rw}}$.
Fig.~\ref{fig:hyper_triplet_counts} reports Test and Tail Accuracy/Macro-F1 on five datasets while varying one triplet count at a time over $\{1,2,3,4\}$.

Overall, DMPRP is reasonably robust to the triplet sampling sizes, but excessively increasing the number of sampled triplets does not necessarily bring further gains.
In most cases, small-to-moderate sampling sizes already provide sufficient structural supervision, while larger values tend to yield saturated improvements or slight fluctuations.

For the node-level triplet number $n_n$, the best performance is typically achieved at a relatively small value.
For example, REDDIT and PROTEINS obtain their strongest Test/Tail results around $n_n=1$, while FRANK performs best near $n_n=2$.
This suggests that node-level relevance signals are highly informative even with limited sampling, and overly increasing node triplets may introduce redundancy rather than additional benefit.

For the subgraph-level triplet number $n_g$, moderate values generally provide a better trade-off between effectiveness and stability.
FRANK and PROTEINS are strongest around $n_g=2$, whereas D\&D remains competitive at slightly larger values.
Overall, these results indicate that subgraph-level supervision complements node-level signals, but does not require aggressive sampling to be effective.

For the random-walk triplet number $n_{\mathrm{rw}}$, the curves show a clearer preference for moderate settings.
Across several datasets, $n_{\mathrm{rw}}=2$ consistently achieves the best or near-best Test and Tail performance, while increasing the number further often leads to saturation or mild degradation.
This trend suggests that a moderate amount of neighborhood-level supervision is sufficient to capture transferable local context, whereas excessive random-walk triplets may introduce noisier or less relevant neighborhoods.

Overall, the results in Fig.~\ref{fig:hyper_triplet_counts} confirm that the effectiveness of multi-granularity triplet mining depends not only on the loss weights, but also on the sampling scale of each view.
In practice, relatively small or moderate triplet counts generally provide a good trade-off between performance and robustness.

\subsection{Scalability Evaluation on Large-scale OGB-MolHIV}
\label{app:ogb_molhiv}

To further evaluate the scalability of DMPRP beyond relatively small TU-style graph classification benchmarks, we conduct additional experiments on the large-scale OGB benchmark \textbf{ogbg-molhiv}~\cite{hu2020open}. 
We adopt the same GIN backbone and follow the size-oriented LTGC protocol by ranking graphs according to the number of nodes and evaluating both the overall test performance and the topology-scarce tail subset.
Since ogbg-molhiv is a binary molecular property prediction dataset, we report ROC-AUC as the evaluation metric.

Table~\ref{tab:ogb_molhiv_additional} shows that DMPRP improves on the GIN backbone on the full test set. The gain is larger on tail molecules, where mean ROC-AUC rises from 0.6170 to 0.7002. This suggests that smaller molecular graphs also benefit from the learned complement on this larger benchmark.

We also examine the role of edge features under heterogeneous structural domains.
Molecular graphs may contain different scaffold-specific structures, and the transferability assumption between head and tail graphs can be weakened when their structural motifs are highly heterogeneous.
To mitigate this issue, we incorporate edge features as complementary relational signals for pattern relevance prediction.
Without edge features, tail performance falls close to that of the GIN baseline.
This result indicates that edge features help DMPRP better capture scaffold-specific structural differences and reduce the risk of negative transfer in heterogeneous molecular graphs.
Nevertheless, we view this as a mitigation mechanism rather than a complete solution, and more effective designs for highly heterogeneous graph domains remain an important direction for future work.

\begin{table}[t]
\centering
\small
\setlength{\tabcolsep}{4pt}
\caption{Additional results on ogbg-molhiv. We report overall test ROC-AUC and tail-subset ROC-AUC.}
\label{tab:ogb_molhiv_additional}
\resizebox{\columnwidth}{!}{\begin{tabular}{@{}lcc}
\toprule
Method / Setting & Test ROC-AUC & Tail ROC-AUC \\
\midrule
GIN & $0.7730 \pm 0.0138$ & $0.6170 \pm 0.0265$ \\
DMPRP & $\mathbf{0.7834 \pm 0.0135}$ & $\mathbf{0.7002 \pm 0.0398}$ \\
\midrule
w/o edge features & $0.7644 \pm 0.0112$ & $0.6184 \pm 0.0323$ \\
w/ edge features  & $\mathbf{0.7834 \pm 0.0135}$ & $\mathbf{0.7002 \pm 0.0398}$ \\
\bottomrule
\end{tabular}}
\end{table}

\subsection{Diagnostic Evidence for the Complement Vector $\mathbf{r}$}
\label{app:r_diagnostic}

The complement $\mathbf{r}=\phi_r(\mathbf{h})$ transforms a graph embedding using patterns learned from head graphs. On PTC, we examine how well $\mathbf{r}$ supports classification on its own, how it changes $\mathbf{h}$ when added to it, and which head graphs it retrieves for tail queries.

\begin{table}[t]
\centering
\footnotesize
\setlength{\tabcolsep}{3pt}
\caption{Diagnostic evidence for the complement vector $\mathbf{r}$ on PTC. For ``only $\mathbf{h}$'' and ``only $\mathbf{r}$'', the results are reported as Accuracy/Macro-F1.}
\label{tab:r_diagnostic}
\resizebox{\columnwidth}{!}{\begin{tabular}{@{}p{0.42\columnwidth}p{0.24\columnwidth}p{0.22\columnwidth}}
\toprule
Evidence & Setting & Result \\
\midrule
only $\mathbf{h}$ & Test & 62.94 / 61.07 \\
only $\mathbf{r}$ & Test & 61.76 / 57.02 \\
only $\mathbf{h}$ & Tail & 60.82 / 60.23 \\
only $\mathbf{r}$ & Tail & 57.96 / 55.83 \\
$\cos(\mathbf{h},\mathbf{r})$ & All / Tail & -0.389 / -0.376 \\
norm ratio of $\mathbf{r}$ to $\mathbf{h}$ & All / Tail & 0.174 / 0.184 \\
Top-1 tail-to-head size gap & using $\mathbf{h}$ / using $\mathbf{r}$ & 30.07 / 21.89 \\
\bottomrule
\end{tabular}}
\end{table}

On tail graphs, a classifier using $\mathbf{r}$ alone achieves 57.96\% accuracy and 55.83\% Macro-F1. Both scores are lower than those obtained using $\mathbf{h}$ alone (Table~\ref{tab:r_diagnostic}). Thus, the learned complement retains signal relevant to the class label.

On both the full set and the tail subset, the residual is about one-fifth of the magnitude of $\mathbf{h}$ on average. Mean cosine similarity is negative in both groups, indicating a tendency for the residual to point partly against $\mathbf{h}$. These measurements describe the size and direction of the adjustment. Quantifying its effect on classification requires comparing predictions made with $\mathbf{h}$ and $\mathbf{h}+\mathbf{r}$ under the same conditions.

For each tail query, we also compare its node count with that of its top-ranked head graph. The mean gap falls from 30.07 nodes when retrieval uses $\mathbf{h}$ to 21.89 when it uses $\mathbf{r}$. The learned representation therefore retrieves head graphs closer in size to the query on average.

\subsection{Qualitative Visualization of Expert Activations}
\label{app:expert_activation}
\begin{figure}[t]
  \centering
  \includegraphics[width=\columnwidth]{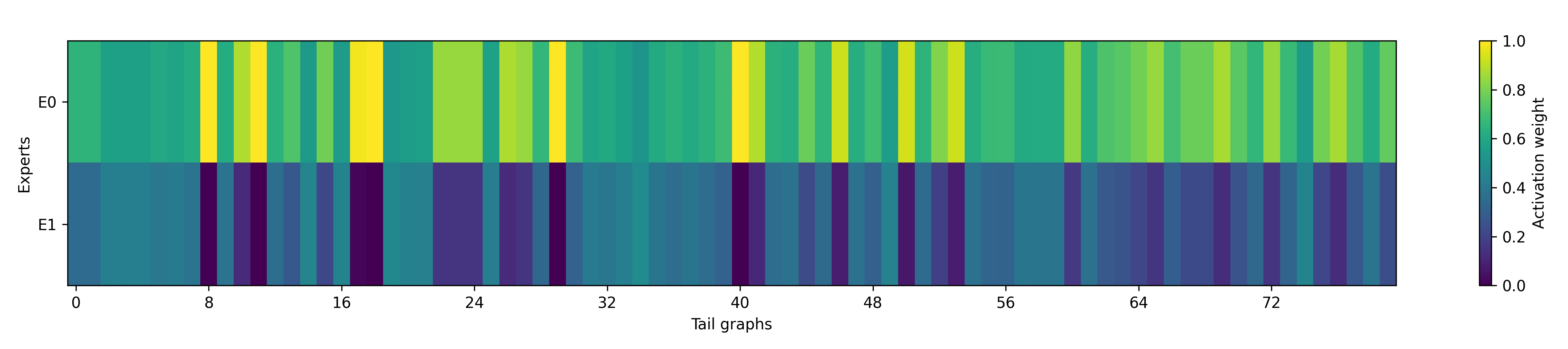}
  \caption{
  Expert activation heatmap on tail graphs.
  Columns denote tail-graph queries and rows denote experts.
  The color indicates the activation weight produced by the query-conditioned gating network.
  Different tail graphs activate the experts with different strengths, showing that DMPRP adaptively selects complementary structural cues.
  }
  \label{fig:expert_activation_heatmap}
\end{figure}

To further illustrate what the dynamic expert memory learns, we provide a qualitative visualization of expert activations on tail graphs.
Fig.~\ref{fig:expert_activation_heatmap} shows the activation weights of different experts for tail-graph queries.
Each column corresponds to one tail graph, and each row corresponds to one expert.
The color intensity indicates the activation weight assigned by the query-conditioned gating mechanism.

As shown in Fig.~\ref{fig:expert_activation_heatmap}, the expert activations vary across tail graphs instead of collapsing to a uniform pattern.
This suggests that the dynamic memory performs query-conditioned expert selection and provides diverse complementary information for different topology-scarce graphs.
Overall, this visualization provides qualitative evidence that the expert-based dynamic memory learns non-trivial and adaptive complement patterns.

\subsection{Extreme-case Evaluation on a Controlled Small-Dense/Large-Sparse Split}
\label{app:ptc_sdls}

To evaluate DMPRP under an extreme topology-conflict scenario, we construct a controlled small-dense versus large-sparse split from the original PTC dataset, denoted as \textbf{PTC-SDLS}.
This experiment is designed as a stress test for the case where graph size and structural density provide different indications of topology richness.
The graph structures, labels, node attributes, and edges are kept unchanged.
We only select two extreme groups according to the conflict score
\begin{equation}
s(G)=z(\bar{d}(G))-z(|V|),
\quad
\bar{d}(G)=\frac{2|E|}{|V|},
\end{equation}
where $z(\cdot)$ denotes standardization over the original PTC dataset. We select the 50 highest-scoring graphs as the small-dense group and the 50 lowest-scoring graphs as the large-sparse group.

\begin{table}[t]
\centering
\small
\caption{Statistics of the controlled PTC-SDLS split. 
$K$ denotes the number of selected head graphs, and Density denotes $2|E|/(|V|(|V|-1))$.}
\label{tab:ptc_sdls_stats}

\resizebox{\columnwidth}{!}{\begin{tabular}{lrrrrrr}
\toprule
\multicolumn{7}{c}{\textbf{Dataset statistics}} \\
\midrule
Dataset & \#Graphs & Avg.$|V|$ & Avg.$|E|$ & \#Cls & \#Attr & $K$ \\
\midrule
PTC-SDLS & 100 & 31.52 & 32.47 & 2 & 16 & 50 \\
\bottomrule
\end{tabular}}

\vspace{0.6em}

\resizebox{\columnwidth}{!}{\begin{tabular}{lrrrr}
\toprule
\multicolumn{5}{c}{\textbf{Group statistics}} \\
\midrule
Group & \#Graphs & Avg.$|V|$ & Avg.$|E|$ & Density \\
\midrule
Small-dense & 50 & 18.70 & 20.08 & 0.121 \\
Large-sparse & 50 & 44.34 & 44.86 & 0.047 \\
\bottomrule
\end{tabular}}

\end{table}

Table~\ref{tab:ptc_sdls_stats} shows how the two groups differ. The small-dense graphs have fewer than half as many nodes on average as the large-sparse graphs, yet their mean edge density is more than twice as high. A size-based partition therefore treats the larger, sparser graphs as head graphs. The conflict score reverses this assignment and lets us examine transfer from smaller, denser graphs to larger, sparser ones.

\begin{table}[t]
\centering
\small
\setlength{\tabcolsep}{4pt}
\caption{PTC-SDLS accuracy (\%, mean $\pm$ standard deviation). DMPRP uses graph size to define the head group; AVG-DEG uses the conflict score $s(G)$.}
\label{tab:ptc_sdls_results}

\resizebox{\columnwidth}{!}{\begin{tabular}{lccc}
\toprule
Method & Test & Head & Tail \\
\midrule
SOLT-GNN 
& $61.00 \pm 8.94$ 
& $72.50 \pm 5.59$ 
& $53.33 \pm 16.24$ \\

DMPRP 
& $64.65 \pm 5.70$ 
& $72.60 \pm 6.85$ 
& $56.67 \pm 6.97$ \\

DMPRP (AVG-DEG) 
& $\mathbf{71.50 \pm 6.12}$ 
& $\mathbf{73.00 \pm 7.14}$ 
& $\mathbf{70.00 \pm 7.07}$ \\
\bottomrule
\end{tabular}}
\end{table}

Table~\ref{tab:ptc_sdls_results} compares the two partition rules. Under the size-based partition, DMPRP improves overall test accuracy from 61.00\% for SOLT-GNN to 64.65\%. DMPRP also improves tail accuracy, while head accuracy is similar for the two methods.

With the average-degree conflict partition, DMPRP (AVG-DEG) reaches 71.50\% overall test accuracy. On the large-sparse graphs now designated as the tail, it achieves 70.00\% accuracy. Because the partition changes which graphs belong to each group, the overall test accuracy provides the direct comparison between the two DMPRP settings. On this controlled split, the higher overall accuracy suggests that smaller graphs can provide transferable structural cues when they are sufficiently dense.

\section{Conclusion}
\label{sec:Conclusion}
\textbf{DMPRP} tackles long-tailed graph classification under \emph{graph-level topology imbalance} by complementing topology-scarce tail graphs with transferable patterns learned from head graphs. It combines a GNN encoder with a \emph{KDEX-style dynamic expert memory} to produce query-conditioned, relevance-aware complements, and leverages multi-granularity triplet mining (node/subgraph/random-walk neighborhoods) for contrastive supervision. Experiments on five size-imbalanced benchmarks show consistent gains in Accuracy and Macro-F1, especially on tail graphs, and ablations confirm the benefits of dynamic memory retrieval and random-walk triplets.


\end{document}